\documentclass[twoside,11pt]{article}

\usepackage{graphicx} 
\usepackage{subfigure} 
\usepackage{graphicx}
\usepackage{subfigure}
\usepackage{epstopdf}
\usepackage{booktabs}
\usepackage{multirow}
\usepackage{url}
\usepackage{color}
\usepackage{hyperref}
\usepackage[noend]{algorithmic}
\usepackage{algorithm}
\usepackage{amsmath}
\usepackage{amssymb}

\definecolor{darkviolet}{RGB}{148,0,211}

\newcommand{\scrD}{\ensuremath{\mathcal{D}}}
\newcommand{\scrL}{\ensuremath{\mathcal{L}}}

\newcommand{\scrX}{\ensuremath{\mathcal{X}}}

\newcommand{\SB}{\ensuremath{\mathbf{S}}}
\newcommand{\DB}{\ensuremath{\mathbf{D}}}

\newcommand{\xiB}{\mbox{\boldmath $\xi$}}
\newcommand{\subsvms}{{\tt SubSVMs}}

\usepackage{jmlr2e}

\jmlrheading{1}{2000}{1-48}{4/00}{10/00}{Srivatsan Laxman, Sushil Mittal and Ramarathnam Venkatesan}

\ShortHeadings{Error Correction in Learning using SVMs}{Srivatsan Laxman, Sushil Mittal and Ramarathnam Venkatesan}
\firstpageno{1}

\begin{document}

\title{Error Correction in Learning using SVMs}

\author{
       \name Srivatsan Laxman \email slaxman@microsoft.com \\
	   \addr Microsoft Research India\\
             ``Vigyan'', \#9, Lavelle Road\\
             Bangalore 560 001, India\\
       \AND
       \name Sushil Mittal \email mittal@stat.columbia.edu \\
       \addr Department of Statistics\\
             Columbia University\\
             New York, NY 10027, USA
       \AND
       \name Ramarathnam Venkatesan \email venkie@microsoft.com \\
       \addr Microsoft Research India\\
             ``Vigyan'', \#9, Lavelle Road\\
             Bangalore 560 001, India\\
       }

\editor{}

\maketitle

\begin{abstract}
This paper is concerned with learning binary classifiers under adversarial label-noise. We introduce the problem of {\em error-correction in learning} where the goal is to recover the original clean data from a label-manipulated version of it, given (i)~ no constraints on the adversary other than an upper-bound on the number of errors, and (ii)~some regularity properties for the original data.  We present a simple and practical error-correction algorithm called \subsvms\ that
learns individual SVMs on several small-size (log-size), class-balanced, random subsets of the data and then reclassifies the training points using a majority vote. Our analysis reveals the need for the two main ingredients of \subsvms, namely class-balanced sampling and subsampled bagging. Experimental results on synthetic as well as benchmark UCI data demonstrate the effectiveness of our approach. In addition to noise-tolerance, $\log$-size subsampled bagging also yields significant run-time benefits over standard SVMs.
\end{abstract}

\section{Introduction}

Learning in the presence of noise is notoriously difficult; there are many negative results regarding hardness of learning under adversarial or malicious noise \cite{BEL03,FGKP06,GR06,hastad97,KSS94,LS11}, while positive results are mostly known only for the case of random noise or under strong distributional assumptions \cite{BFKV96,KKMS08,SNM10,servedio03}. Somewhat more encouraging results exist in max-margin settings \cite{BS00,HRZ06,SSS10,XCS06} but these methods are computationally prohibitive even for reasonably-sized data. 

In this paper, we investigate the learning of binary classifiers under adversarial (worst-case) label-noise. We introduce the problem of {\em error-correction} in learning, as the task of correcting the label-errors in training data, $\widehat{\DB}$, given that the original (clean) data, $\DB$, intrinsically satisfies some regularity properties. (Given negative results such as \cite{GR06} regarding the hardness of learning better-than-random hyperplanes even from
nearly-separable data, some notion of regularity becomes essential). Informally, $\DB$ is said to be $r$-regular if SVMs trained on very small random $r$-subsets  of $\DB$, make less than $\theta$-fraction errors over all of $\DB$.  We show that every linearly separable $\DB$  exhibits some regularity, and that such a $\DB$ can be recovered from any $\widehat{\DB}$ with roughly $(\frac{1}{2}-2\theta-O(\log^2r))$-fraction of errors. The main idea in our analysis is to apply
margin-based generalization bounds under a chosen sampling distribution over $\DB$ and to then adjust the bounds for the noise in $\widehat{\DB}$. To the best of our knowledge, this is the first positive result that is known about learning classifiers under adversarial label-errors.



Our algorithm for error-correction, called {\subsvms} ({\tt Sub}sample bagging of {\tt SVMs}) is as follows: Train SVMs on suitably-small, class-balanced, random subsets of $\widehat{\DB}$ and reclassify every training point using a simple majority vote. We show that class-balanced sampling over $\widehat{\DB}$ minimizes the worst-case probability of drawing less than any-chosen-number of clean points per class from $\widehat{\DB}$. The number of worst-case errors  that each SVM in the ensemble makes can grow as the squared-log of the subsample-size used, and this leads us to the final error-correction performance of \subsvms.

In experimental work, we first study the error-correction achievable on synthetic linearly separable data. By comparing against performance under uniform sampling (common in standard bagging) we show that class-balanced sampling plays a vital role in error-correction. Then we show that error-correction based on {\subsvms} leads to better classifiers which outperform regular SVMs on a range of benchmark data sets from the UCI Machine Learning Repository. Our experiments also
clearly demonstrate superiority of {\subsvms} over regular bagging. We inject high-levels of label-noise in the training data sets (Number of errors was fixed at 75\% of the size of the minority class). On previously unseen (clean) test sets, {\subsvms} even outperformed SVMs that directly used the full test sets for cross-validation. Subsampling at logarithmic sizes also gives {\subsvms} substantial run-time advantages over standard
SVMs and regular bagging.

{\bf Related Work:} Several results show that learning under adversarial noise can be NP-hard \cite{hastad97,KSS94,FGKP06,GR06}. Better results (polynomial-time algorithms) are known in the context of learning max-margin classifiers from noisy data \cite{HRZ06,SSS10,XCS06}. However, these techniques are computationally prohibitive in practice, e.g., the method proposed in \cite{XCS06} uses SDP solvers that can become impractical even for a hundred training points. Many boosting algorithms, with convex potential functions, have also been shown vulnerable to random classification noise \cite{LS10}.

In statistical (rather than adversarial) settings, generalization results for SVMs demonstrate efficient learnability when training and test points are drawn {\em iid} from the same (even if noisy) distribution \cite{CS00}.  Some works have focused on the ineffectiveness of SVMs in the presence of outliers and for noisy class-imbalanced data (e.g., see \cite{AKJ04,TG05,NB07}), albeit without formal analysis. Recently, large-margin half-spaces were shown to be efficiently learnable under small amounts of malicious noise \cite{LS11}. Similarly, \cite{DS09} demonstrates learning from multi-teacher data, where a small number of teachers can replace randomly chosen examples arbitrarily. A general framework for distribution-dependent learning in-the-limit was proposed in \cite{CM08}; the focus, however, was on establishing informational limits rather than sample complexities.  We consider learning under adversarial label-errors given that
the original data satisfies some regularity properties. Our error-model is relevant both when the label-errors are inadvertent, whether systematic or random, and when errors are introduced by an adversary explicitly trying to mislead the learning process.

Several studies investigated why (and under what conditions) bagging works by formalizing different notions of stability for predictors and by showing that bagging reduces the variance of unstable predictors (see, e.g., \cite{breiman96b, BY02, EEP05,grandvalet04}). Experimental bias-variance analysis of random aggregation and bagging of SVMs demonstrated that working with small samples achieves greater reduction in the variance component of error than standard bagging
(see \cite{valentini04}). In another related work, \cite{BF99} presented an experimental study of various methods for identifying mislabeled data. All these studies, including the ones that analyze bagging, restricted attention to distribution-based models, rather than adversarial settings.

\section{Error correction problem in learning}
\label{sec:ecpil}

Let $\DB = \{(x_i, y_i):i = 1, \ldots, \ell\}$ be the set of examples in a binary classification problem; the feature vectors, $x_i$, come from some domain $\scrX$ and the class-labels, $y_i$, take values from $\{-1,+1\}$. The proportion of minority class points in $\DB$ is denoted $\beta,\ 0<\beta\leq 0.5$.

Let $\Psi_\DB$ denote a binary SVM classifier trained on $\DB$\footnote{$\Psi_\SB$ denotes the SVM trained on $\SB$, etc.}; for $x\in\scrX$, the classifier returns the label  $\Psi_\DB(x)\in\{-1,+1\}$. We assume that $\Psi_\DB$ is suitable for the given classification task. However, $\DB$ is not available to train the learning algorithm. Instead, the learner only has access to $\widehat{\DB} = \{(x_i,\widehat{y}_i):i = 1, \ldots, \ell\}$, which is a {\em label-manipulated}
version of $\DB$\footnote{$\widehat{\DB}$ is also referred to as the {\em corrupted} or {\em noisy} data.}.


The adversary is allowed to flip labels of {\em no more than} $\rho\beta\ell$ examples in $\DB$, where $\rho$ is referred to as the {\em error parameter}. Since we place no other restrictions on the points the adversary can manipulate, we must have the constraint $0\leq\rho<1$ (otherwise, we may be left with no training examples for one class).


The error-correction problem is concerned with recovering the original clean data $\DB$ (or a close approximation of it) from its label-manipulated version $\widehat{\DB}$. To this end, we will allow some `regularity' assumptions on the original data $\DB$, which essentially guarantee that SVMs trained on sufficiently-small random subsets of $\DB$ can classify the points in $\DB$ with high accuracy. Regularity is an intrinsic property of the original data, which can manifest and be measured in many ways; one way is to measure the redundancy structure exposed by the quadratic program underlying the max-margin formulation of SVMs. 

\begin{definition}
[Data Regularity] Let $\scrD_\ast$ be any (discrete) probability distribution over $\DB$ and let $\SB\sim\scrD_\ast$, $|\SB|\geq r$, denote a collection of points drawn {\em iid} from $\scrD_\ast$. For any $\delta<0.5$ and $\theta<0.5$, $\DB$ is said to be {\em $r$-regular} at $(\delta,\theta)$ if with probability at least $1-\delta$ over choice of $\SB$, the expected error-rate of $\Psi_\SB$ does not $\theta$ with respect to test examples also drawn {\em iid} from $\scrD_\ast$.
\label{def:regularity}
\end{definition}


We are interested in regularity at small $r$, such as at $O(\log\ell)$ or $O(\log^2\ell)$. Data regularity can be thought of as a measure of {\em redundancy} needed to admit learning in the presence of adversarial label-noise. This is, in a sense, akin to the redundancy encoded into a message for enabling error-correction in coding theory. Regularity is a simple property that is satisfied by data from which good binary classifiers can be easily learnt, e.g., every linearly separable data set is regular.
\begin{lemma}
[Separability implies Regularity] Consider any linearly separable $\DB$ with margin $\gamma$. For any fixed $\delta<0.5$ and $\theta<0.5$, there exists $r\in\mathbb{Z}^+$ such that $\DB$ is $r$-regular at $(\delta,\theta)$.
\label{lem:separableRegular}
\end{lemma}
The proof makes use of the following 2-norm soft-margin bound from SVM generalization theory \cite{CS00}:
\begin{theorem}
\citep[Theorem 4.22]{CS00} Consider thresholding real-valued linear functions $\scrL$ with unit weight vectors on an inner product space $\scrX$ and fix $\gamma\in\mathbb{R}^+$. There is a constant $c$, such that for any probability distribution $\scrD$ on $\scrX\times\{-1,+1\}$ with support in a ball of radius $R$ around the origin, with probability $1-\delta$ over $\ell$ random (training) examples $\DB=\{(x_1,y_1),\ldots,(x_\ell,y_\ell)\}$, any hypothesis $f\in\scrL$ has error no more than
\begin{equation}
\Pr_{(x,y)\sim\scrD}[ f(x) \neq y ] \leq \frac{c}{\ell}\left( \frac{R^2+\lVert{\xiB}\rVert_{2}^2}{\gamma^2}\log^2\ell+\log\frac{1}{\delta}\right)
\label{eq:2normBound}
\end{equation}
where $\xiB=(\xi_1,\ldots,\xi_\ell)$ is the margin slack vector with respect to $f$ and $\gamma$. The entries of $\xiB$ are fixed as follows: $\xi_i=\max(0,\gamma-y_i f(x_i))$, $i=1,\ldots,\ell$.
\label{thm:2normGE}
\end{theorem}
Since $\DB$ is separable with margin $\gamma$, every subset of $\DB$ is also separable with margin at least $\gamma$. Thus, the max-margin separator of every subset of $\DB$ will have margin slack vector $\xiB=0$ (with respect to the chosen subset). Fixing $\scrD=\scrD_\ast$ in {\em Theorem~\ref{thm:2normGE}}, the generalization error of $\Psi_\SB$ trained on any $\SB\sim \scrD_\ast$,  $|\SB| = r$, is given by
\begin{equation}
\Pr_{(x,y)\sim\scrD_\ast}[ \Psi_\SB(x) \neq y ] \leq \frac{c}{r}\left( \frac{R^2}{\gamma^2}\log^2 r+\log\frac{1}{\delta}\right)
\label{eq:thetaGE}
\end{equation}
{\em Lemma~\ref{lem:separableRegular}} follows since the RHS of (\ref{eq:thetaGE}) is $O(\log^2 r/r)$. 

\begin{definition}
[Error-correction in Learning] Given that $\DB$ and $\widehat{\DB}$ disagree on no more than $\rho\beta$-fraction of labels, and given that $\DB$ satisfies some regularity properties, the problem of {\em error-correction in learning} is to recover a data set $\widetilde{\DB}$ with as few label disagreements with $\DB$ as possible.
\label{def:ecproblem}
\label{def:ecsvm}
\end{definition}
We make no assumptions regarding the nature of label-errors (such as if they are statistical or otherwise), or regarding the separate values of error-parameter ($\rho$) and true fraction of minority-class ($\beta$); we are only given that the total fraction of label-errors does not exceed $\rho\beta$, $0\leq\rho<1$ and $0<\beta\leq0.5$.

\section{The {\subsvms} algorithm}
\label{sec:sbsvm}

\begin{algorithm}[t]
\caption{{\tt [SubSVMs]} Subsampled bagging of SVMs}
\label{algo-csbsvm}
\label{algo-sbsvm}
\begin{algorithmic}[0]
\REQUIRE Corrupted data $\widehat{\DB} = \{ (x_1,\widehat{y}_1),\ldots,(x_1,\widehat{y}_\ell)\}$; size, $s$, of subsample; sampling bias $p$; number of SVMs $J$ (typically, $p=\frac{1}{2}$ and $s=\log \ell$ or $s=\log^2 \ell$)
\ENSURE Error-corrected data $\widetilde{\DB} = \{(x_1,\widetilde{y}_1),\ldots,(x_1,\widetilde{y}_\ell)\}$

\STATE   
\STATE $/\ast$ Training $\ast/$ 
\FOR{$j=1$ to $J$}
	\STATE Draw random subset $\widehat{\SB}_j\sim\scrD_{p\widehat{\DB}}$ of size $|\widehat{\SB}_j|=s$
	\STATE Train SVM $\Psi_{\widehat{\SB}_j}$
\ENDFOR
\STATE

\STATE $/\ast$ Error-correction $\ast/$
\FOR{$i=1$ to $\ell$}
\STATE Set $\widetilde{y_i}$ to the majority label 
in $\{\Psi_{\widehat{\SB}_1}(x_i),\ldots,\Psi_{\widehat{\SB}_J}(x_i)\}$
\ENDFOR

\STATE Output $\widetilde{\DB} = \{(x_1,\widetilde{y}_1),\ldots,(x_1,\widetilde{y}_\ell)\}$
\end{algorithmic}
\end{algorithm}

We first define a key ingredient of the {\subsvms} algorithm that we refer to as {\em $p$-biased sampling}.
\begin{definition}
[$p$-biased Sampling] The process of {\em $p$-biased sampling} of $\widehat{\DB}$ refers to the following two steps, executed in the stated order: (1)~choose the minority class\footnote{If both classes of $\widehat{\DB}$ are of identical size, one of them is arbitrarily fixed as the `minority class'.} of $\widehat{\DB}$ with probability $p$ (or the other class with probability $1-p$) and (2)~pick a point uniformly at random from the restriction of $\widehat{\DB}$ to the chosen class. The corresponding sampling distribution is denoted $\scrD_{p\widehat{\DB}}$ and $\widehat{\SB}\sim\scrD_{p\widehat{\DB}}$ denotes that $\widehat{\SB}$ is a random collection of points drawn {\em iid} with respect to $\scrD_{p\widehat{\DB}}$.
\label{def:cbSampling}
\end{definition}

The case of $p=0.5$ is referred to as {\em class-balanced sampling} of $\widehat{\DB}$; if $\widehat{\beta}$ denotes the fraction of minority class points in $\widehat{\DB}$, the case of $p=\widehat{\beta}$ is equivalent to {\em uniform sampling} over $\widehat{\DB}$.

{\em Algorithm~\ref{algo-sbsvm}} lists the pseudo-code for {\em subsampled bagging of SVMs} ({\subsvms}).
Our analysis (in Secs.~\ref{sec:ec-sbsvm-theory}-\ref{sec:importance}) reveals two important aspects of {\subsvms}:

\begin{itemize}
\item Class-balanced sampling provides optimal protection against worst-case label-errors.
\item The fraction of errors that can be tolerated ($\rho\beta$) reduces as the squared-log of sample-size $s$.
\end{itemize}
Based on the above, we use class-balanced sampling ($p=1/2$) and choose $s$ to be $\log \ell$ or $\log^2 \ell$.

\subsection{Error correction analysis}
\label{sec:ec-sbsvm-theory}

Our analysis uses the margin-based generalization bound for SVMs with respect to a sampling distribution over the original (clean) data $\DB$ and then adjusts the bound to accommodate the number of label-errors in the corrupted training set $\widehat{\DB}$.

Consider the general case of {\em Algorithm~\ref{algo-sbsvm}}, where the random subsets $\widehat{\SB}_j$ are drawn {\em iid} from $\scrD_{p\widehat{\DB}}$. Let $\DB$ be linearly separable with margin
$\gamma$. Consider a set of points $\widehat{\SB} \sim \scrD_{p\widehat{\DB}}$. We now need to compute the expected error-rate of $\Psi_{\widehat{\SB}}$ with respect to test points drawn uniformly from $\DB$ (This is the main quantity of interest in the error-correction setting). For this, we first compute the expected error-rate $\epsilon$ when the training and test cases are both drawn {\em iid} from $\scrD_{p\widehat{\DB}}$. This is done by using {\em Theorem~\ref{thm:2normGE}}
\citep[Theorem~4.22]{CS00} with $f=\Psi_{\widehat{\SB}}$ and $\scrD=\scrD_{p\widehat{\DB}}$ (See next paragraph for details). The error-rate can at most become $\epsilon/p^\ast$, where $p^\ast=\min\{p,1-p\}$, when considering test cases drawn {\em uniformly} from $\widehat{\DB}$\footnote{See Appendix~\ref{sec:proof-error-rate} for a short proof.}. Finally, in any uniformly drawn sample from $\widehat{\DB}$, the expected fraction of label disagreements with respect to the corresponding points in $\DB$ is $\rho\beta$. Hence, the desired expected error-rate of $\Psi_{\widehat{\SB}}$, where $\widehat{\SB}\sim\scrD_{p\widehat{\DB}}$ but the test points are drawn uniformly from $\DB$, is given by $\epsilon/p^\ast+\rho\beta$.

We now return to the computation of error-rate $\epsilon$ when train and test points are both drawn {\em iid} from $\scrD_{p\widehat{\DB}}$. Whenever $\widehat{\SB}$ contains {\em at least} $r/2$ clean points per class, the SVM of the corresponding $r$-size (clean) subset of $\widehat{\SB}$ would make no more than $(s-r)$ mistakes on the rest of $\widehat{\SB}$. Each of these mistakes would be no farther than $2R$ from either supporting hyperplane. Also, the
margin of this SVM would be at least $\gamma$ (the max-margin achieved on the whole of $\DB$). The 2-norm SVM objective has the same form as the error-bound in (\ref{eq:2normBound}). Hence, we apply {\em Theorem~\ref{thm:2normGE}} with $\lVert \xiB \rVert^2_2=4R^2(s-r)$ and with margin $\gamma$, to obtain the generalization bound, $\epsilon$.
If $\eta$ is an upper-bound on the probability that $\widehat{\SB}$ contains {\em less than} $r/2$ clean points from either class, then with probability at least $(1-\eta-\delta)$
\begin{equation}
\Pr_{(x,y)\sim\scrD_{p\widehat{\DB}}}[\Psi_{\widehat{\SB}}(x) \neq y ] \leq  \frac{c}{s}\left( \frac{R^2+4R^2(s-r)}{\gamma^2}\log^2s+\log\frac{1}{\delta}\right) \stackrel{\rm def}{=} \epsilon. \nonumber
\label{eq:epsilon}
\end{equation}
Recall that this error-rate, $\epsilon$, over test points drawn from $\scrD_{p\widehat{\DB}}$, translates to an error-rate of $\epsilon/p^\ast+\rho\beta$ for test points drawn uniformly over $\DB$. Thus, the final expression for probability of error of $\Psi_{\widehat{\SB}}$ with respect to test points drawn uniformly from $\DB$, denoted $\varphi$, can be written as follows:
\begin{equation*}
\Pr[\Psi_{\widehat{\SB}}(x)\neq y] \leq (1-\eta-\delta)\left[\frac{\epsilon}{p^\ast}+\rho\beta\right] + \eta+\delta \stackrel{\rm def}{=} \varphi.
\label{eq:finalProbError}
\end{equation*}
We use $J$ SVMs based on $J$ random sets such as $\widehat{\SB}$. Thus, if $\varphi<0.5$, then (by Hoeffding Inequality \cite{hoeffding63}) the probability of a majority vote making a mistake with respect to $\DB$ cannot exceed $\exp[-2J(0.5-\varphi)^2]$. This gives us error-correction (in the sense that $\DB$ can be correctly recovered from $\widehat{\DB}$). To enforce the condition $\varphi<0.5$, we must have $\rho\beta < 1-\epsilon/p^\ast-[2(1-\eta-\delta)]^{-1}$. Finally, if $\DB$ is $r$-regular at $(\delta,\theta)$, then we have
\begin{align}
\epsilon = & \frac{c}{s}\left( \frac{R^2}{\gamma^2}\log^2s+\log\frac{1}{\delta}\right) + \frac{c}{s}\left(\frac{4R^2(s-r)}{\gamma^2}\log^2s\right) \nonumber  \\ \leq & ~~\theta + \frac{c}{s}\left(\frac{4R^2(s-r)}{\gamma^2}\log^2s\right).
\end{align}
This leads to our main result about \subsvms: 
\begin{theorem}
[Error-correction]
Consider linearly separable $\DB$ with margin $\gamma$ and $\beta$-fraction of minority-class points. Fix $\delta < 0.5$ and let $\DB$ be $r$-regular at $(\delta,\theta)$. Consider $\widehat{\DB}$ with error-rate $\rho$ and $\widehat{\SB}\sim \scrD_{p\widehat{\DB}}$, $|\widehat{\SB}|=s$. Let $\Pr[\widehat{\SB}{\rm \ contains\ } < r/2 {\rm \ clean\ points\ per\ class}] \leq \eta$.
If the number of label-errors in $\widehat{\DB}$ is bounded by
\begin{equation}
\rho\beta < 1-2\theta - \left[\frac{1}{2(1-\eta-\delta)}  + \frac{4R^2c(s-r)\log^2s}{\gamma^2s}\right]
\label{eq:rhobeta}
\end{equation}
where $R$ denotes the radius of the ball enclosing the data and $c$ is the constant from {\em Theorem~\ref{thm:2normGE}}, then the probability of error for \subsvms\ with respect to points drawn uniformly from $\DB$ is at most $\exp\left[-2J(0.5 - \varphi)^2\right]$, where $\varphi= \eta+\delta + (1-\eta-\delta)\left[\epsilon/p^\ast+\rho\beta\right]$ and $p^\ast=\min\{p,1-p\}$.
\label{thm:ec-subsvm}
\end{theorem}
Hence, perfect error-correction is attained for $\varphi<0.5$.

\subsection{Importance of Class-balanced Sampling}
\label{sec:importance}

\begin{figure}[t]
\centering
\includegraphics[width = 8cm]{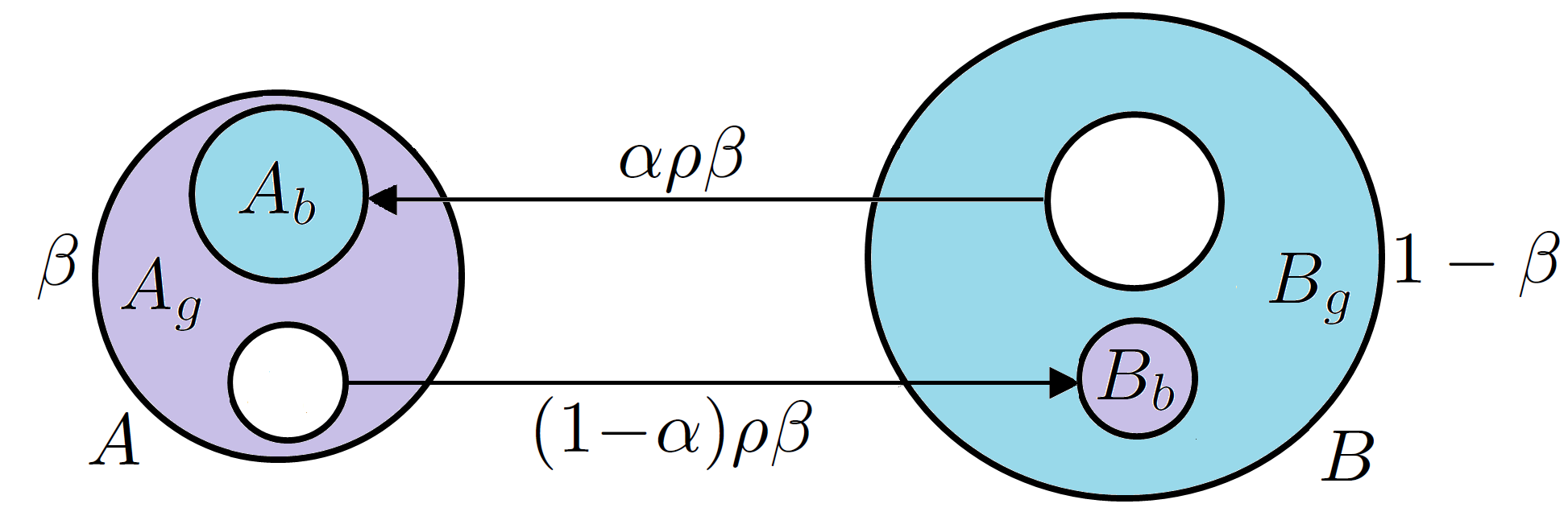} \\

\caption{The data corruption process. Let $A$ be the minority class with $\beta$-fraction of points in $\DB$. $A_b$ represents the $\alpha$-fraction of corrupted points, originally in class-$B$, but wrongly assigned to class-$A$ in $\widehat{\DB}$; similarly $B_b$ represents the class-$A$ points in $\DB$ that were mislabeled as class-$B$ in $\widehat{\DB}$. A total of $\rho\beta$-fraction of points are corrupted in $\widehat{\DB}$.}

\label{fig:corruption}
\end{figure}

The bound in  (\ref{eq:rhobeta}) has two groups of parameters. In the first group, we have $r$, $\delta$ and $\theta$, which are fixed by the regularity properties of $\DB$. In the second group, we have $s$ and $\eta$, which are both determined by our sampling strategy. Since $\eta$ depends on the sampling bias $p$, we now discuss how to fix $p$ and $s$ for optimal error-correction performance.

From (\ref{eq:rhobeta}) it is clear that, to maximize the number of errors that can be tolerated, we must minimize the quantity in square brackets. The first term inside the brackets is minimized when $\eta$ is minimum. Fig.~\ref{fig:corruption} provides a graphical depiction of the data corruption process. The optimal value of $\eta$ typically depends on the direction-of-attack parameter, $\alpha$, the error parameter $\rho$, and the true size, $\beta$, of the minority class in $\DB$. However, neither of these is known to the learner; only an upper-bound on the fraction of label-errors in $\widehat{\DB}$ is known. So we design our algorithm to limit the impact of worst-case label-errors. Specifically, we choose $p=0.5$ since it minimizes $\eta$ in a manner that is agnostic to the true values of $\alpha$, $\rho$ and $\beta$. We state this formally in {\em Lemma~\ref{lem:cbsampling}} below.

\begin{lemma}
[Class-balanced Sampling] Fix any $r\in\mathbb{Z}^+$. Given $\DB$ with $\beta$-fraction of minority-class points and $\widehat{\DB}$ with at most $\rho\beta$-fraction label-errors w.r.t. $\DB$, class-balanced sampling of $\widehat{\DB}$ minimizes a worst-case upper-bound on $\eta$ (probability that the sample drawn contains less than $r/2$ clean points per class) if the size, $s$ ($\geq r$), of the sample satisfies
\label{lem:cbsampling}
\begin{equation}
s \geq 2r + 4\left(r\log2+\log^22-\log4\right)^\frac{1}{2} + \log16 - 4
\label{eq:smin}
\end{equation}
\end{lemma}

The main intuition behind the proof is that, in the absence of any specific information regarding $\rho$, $\beta$ and $\alpha$, choosing the sampling bias $p$ on either side of $0.5$ is vulnerable to one of the attack directions, thereby increasing the worst-case value of $\eta$. (See Appendix~\ref{sec:proof-optimality-cb} for the proof).

The second term inside the square brackets of (\ref{eq:rhobeta}) is smallest (and equal to zero) for $s=r$. However, {\em Lemma~\ref{lem:cbsampling}} shows that this is not optimal for $\eta$, since $s=r$ fails the condition in (\ref{eq:smin}). In fact, for smaller $s$, $\eta$ may even be maximized at $p=0.5$; in general, the minimizer of $\eta$ will no longer be agnostic to $\rho$, $\beta$ and $\alpha$. However, when $s$ is set to the lower-bound of (\ref{eq:smin}), the second term inside square brackets of (\ref{eq:rhobeta}) becomes $O(\log^2r)$. This gives us our next lemma.
\begin{lemma}
[Subsampled Bagging] Let $\DB$ be linearly separable and $r$-regular at $(\delta,\theta)$ and let $\widehat{\DB}$ contain at most $(\rho\beta)$-fraction of adversarial label-errors. {\subsvms} based on class-balanced sampling and with subsample-size, $s$, set to the lower-bound in (\ref{eq:smin}), can perfectly recover the original $\DB$, provided the fraction of label-errors in $\widehat{\DB}$ is bounded above as follows:
\begin{equation}
\rho\beta < 1-2\theta - \left[\frac{1}{2(1-\eta-\delta)} + O(\log^2r)\right]
\label{eq:rhobetalemma}
\end{equation}
\label{lem:subbagging}
\end{lemma}

Since the above lemma requires $s$ to be set at the lower-bound of (\ref{eq:smin}) it might appear that we are operating on a knife-edge for choosing the subsample size. Luckily, this is not the case, because if the data is regular at $r$, it would also be regular with same $\theta$ for every $r'>r$. Hence, we could set $s$ to the lower bound in (\ref{eq:smin}) corresponding to $r'$ and the above Lemma would still hold, though with $O(\log^2r')$ rather than $O(\log^2r)$ inside the square brackets. As a result, the number of worst-case errors allowed reduces for $r'>r$ and this is the reason why we use {\em subsampled} bagging. Typically, we choose $s$ to be $\log\ell$ or $\log^2\ell$ (rather than $\ell$, which is the usual case in bagging). As long as the data is $r$-regular for some $r<s$ that satisfies (\ref{eq:smin}) \subsvms\ will give us error-correction. As a side-benefit subsampling at logarithmic sizes will give us dramatic run-time advantages over regular SVMs. Our experimental results clearly demonstrate this aspect of \subsvms.

\section{Experiments}
\label{sec:experiments}
We present experimental results of \subsvms\ on simulated, linearly separable data as well as \texttt{LIBSVM} extracts of some UCI data sets\footnote{\small http://www.csie.ntu.edu.tw/$\sim$cjlin/libsvmtools/datasets}. SVMs are known to perform well on these data sets, so they can play the role of clean data in our experiments. 

Our data corruption process follows Fig. \ref{fig:corruption}. Given `clean' training data $\mathbf{D}$ of size $\ell$ with minority class of size $\beta \ell, 0 < \beta \leq 0.5$, the parameters $\rho$ and $\alpha$ control the corruption. We randomly pick $\rho\beta\ell$ points for corruption, of which, $\alpha$-fraction are picked uniformly at random from the minority class and $(1-\alpha)$-fraction from the other. By varying the \textit{attack direction} $\alpha$, we generated a wide range of corrupted data with different degrees of difficulty for binary classification.

\begin{figure*}
\centering
\begin{tabular}{@{\hspace{-0.2cm}}c@{\hspace{-0.1cm}}c@{\hspace{-0.1cm}}}
\includegraphics[width = 6cm]{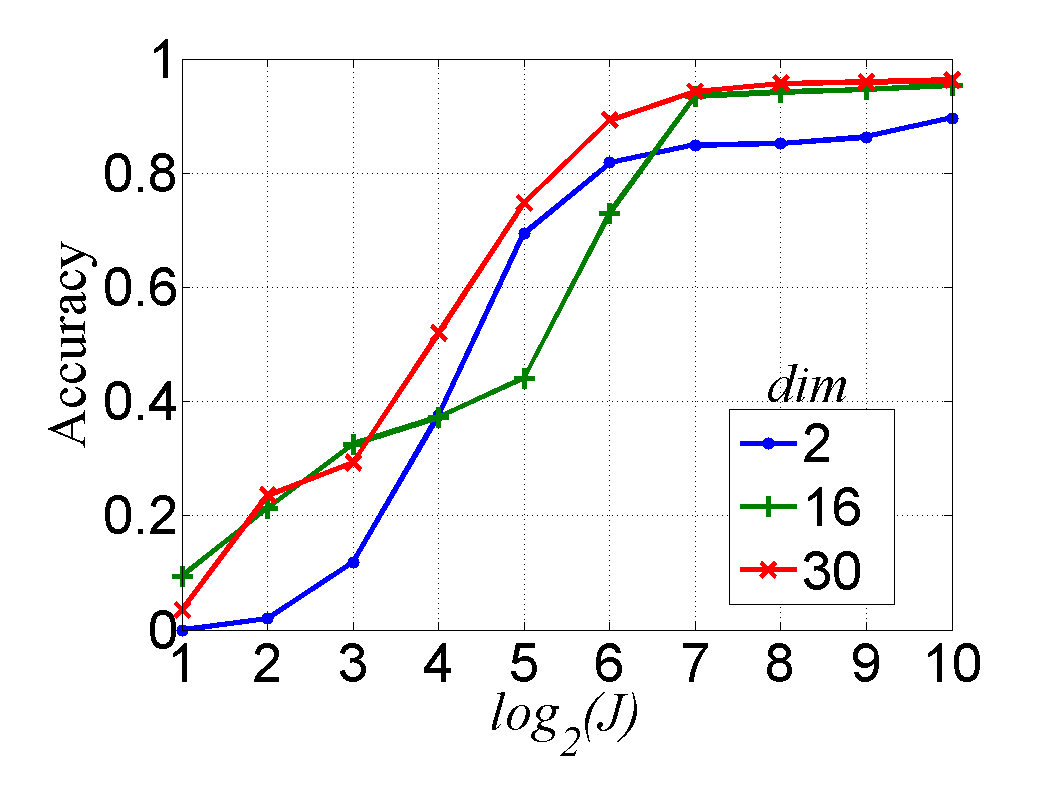} &
\includegraphics[width = 6cm]{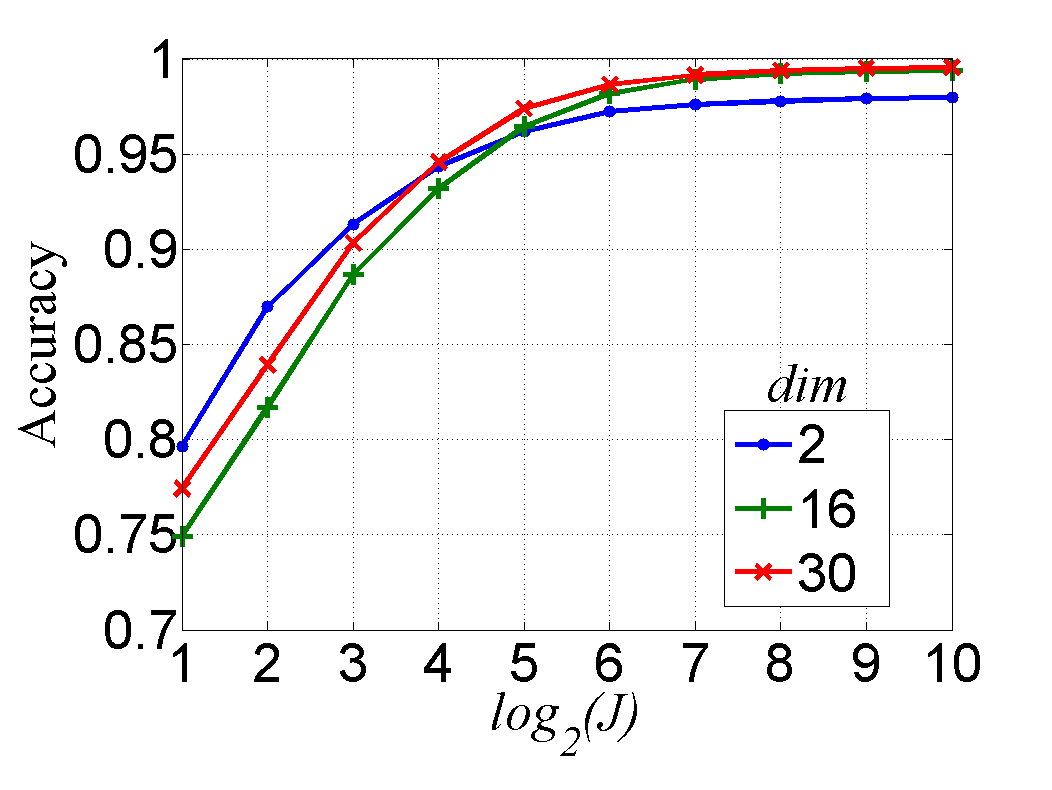} \\
(a) & (b) \\
\includegraphics[width = 6cm]{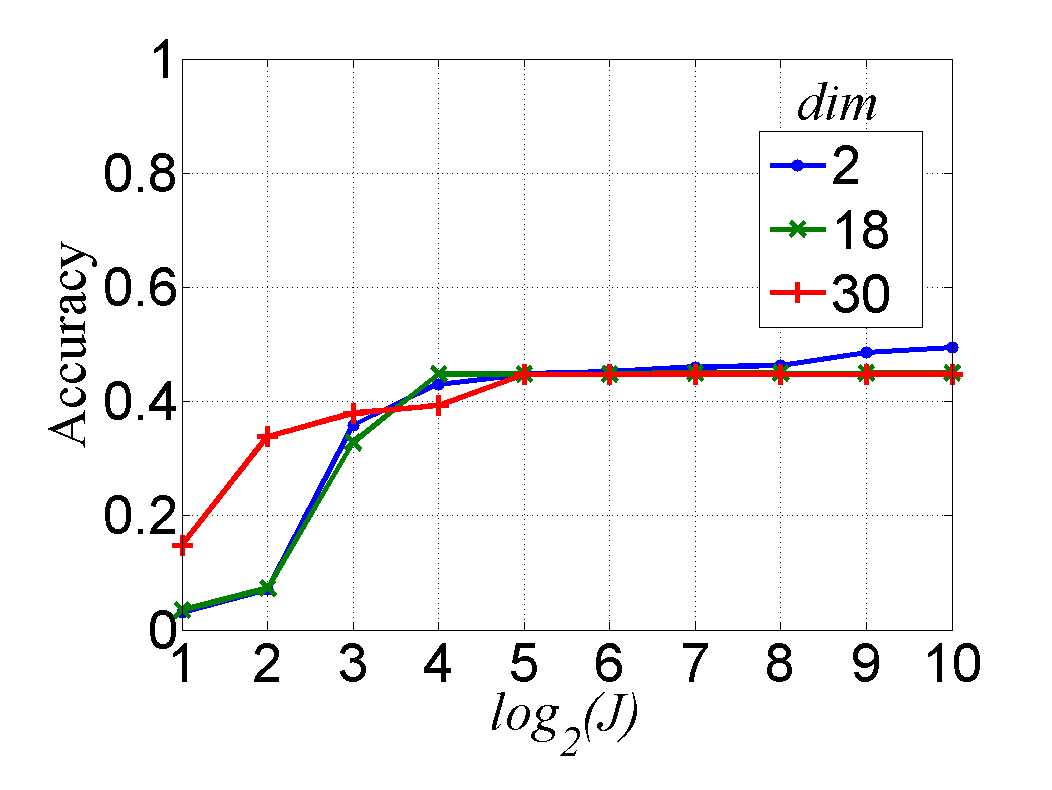} &
\includegraphics[width = 6cm]{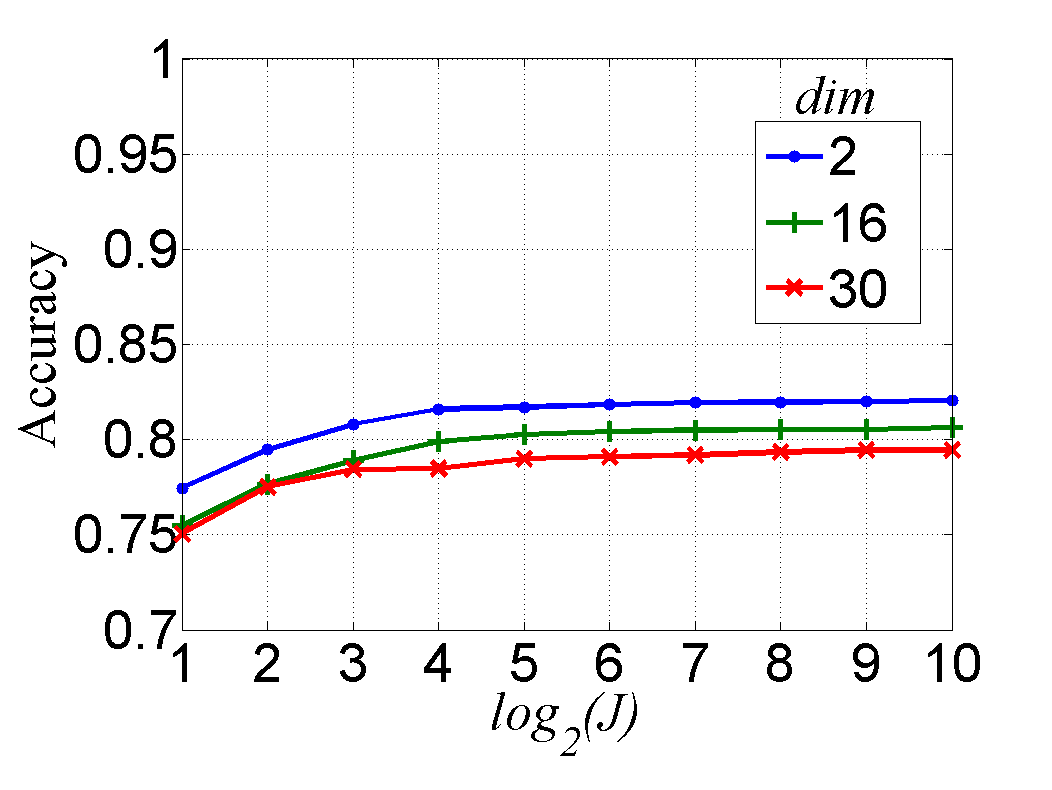} \\
(c) & (d) \\
\end{tabular}
\caption{Importance of class-balanced sampling $(p=1/2)$ in \textit{Algorithm~\ref{algo-sbsvm}}. (a) Worst-case and (b) average-case accuracy obtained using class-balanced sampling $(p=1/2)$ over different label-manipulated data sets as a function of $J$. (c) Worst-case and (d) average-case accuracy obtained using \subsvms~ with uniform sampling $(p=\beta)$.}
\label{fig:syn-expt1}

\end{figure*}

\subsection{Synthetic Data Experiments}

In the first experiment, we generated `clean' data sets $\mathbf{D}$ comprising of 1000 $d$-dimensional data points from a mixture of two Gaussian distributions, each with a covariance of $0.1\mathbf{I}_{d}$ and a distance of two units between means. Three values of $d$ were used: 2, 16 and 30. A constant margin of 0.2 units was enforced and misclassified points were manually removed. The value of $\beta$ was varied between [0.05,~0.5] in steps of 0.05, $\rho=0.75$ and $\alpha$ was varied between [0.0,~1.0] in steps of 0.25.

We studied the importance of class-balanced sampling in \textit{Algorithm~\ref{algo-sbsvm}} (\subsvms) by comparing two versions of it - one with class-balanced sampling $(p=1/2)$ and the other with uniform sampling $(p=\beta)$. For every $d$, the data corresponding to each $[\beta,\alpha]$ pair was subjected to 10 random corruptions. Figs. \ref{fig:syn-expt1}a and \ref{fig:syn-expt1}b summarize the results for class-balanced sampling and Figs. \ref{fig:syn-expt1}c and \ref{fig:syn-expt1}d  for  uniform sampling. As expected, based on Theorem \ref{thm:ec-subsvm}, the number of mistakes made decays exponentially with increasing $J$. Near-perfect error-correction is achieved using $p=1/2$ for $J$ as small as $2^7$. For $p=\beta$, the worst-case and average-case performances are worse by about $60\%$ and $20\%$, respectively. This experimentally validates Lemma \ref{lem:cbsampling} for using class-balanced sampling in \subsvms.

\subsection{UCI Data Experiments}
\label{sec:UCIData}
We now report the performance of \subsvms\ on held-out test data using the LIBSVM UCI extracts. There can be two ways to test this, either the error-corrected training data can be used to retrain a fresh standard SVM or we can just use majority voting over the $J$ SVMs already trained in \subsvms. In our experiments, both these approaches yielded very similar results. Therefore, we avoid retraining cost and report results using the majority voting method.

\begin{table*}[!t]
\centering
\begin{tabular}{|c|c|c|c|c|c|}
\hline
		&				& 	\multicolumn{2}{c|}{Training set}		& 	\multicolumn{2}{c|}{Test set} 	\\ \cline{3-6}
Data	&	Feature		& Total 		& Size of			& Total	& Size of \\
set		&	dimension	& size			& minority class	& size	& minority class \\ \hline
a1a		& 				& 1605			& 395 (25\%) 						& 30956 		& 7446 (24\%) \\
a2a		& 				& 2265			& 572 (25\%) 						& 30296 		& 7269 (24\%) \\
a3a		& 				& 3185			& 773 (24\%) 						& 29376 		& 7068 (24\%) \\
a4a		& 				& 4781			& 1188 (25\%) 						& 27780 		& 6653 (24\%) \\
a5a		& 	123			& 6414			& 1569 (25\%) 						& 26147 		& 6272 (24\%) \\
splice	& 	60			& 1000			& 483 (48\%) 						&  2175 		& 1044 (48\%) \\
mushrooms&	112			& 6093			& 2937 (48\%) 						&  2031 		& 979 (48\%) \\
svmguide1&	4			& 3089			& 1089 (35\%) 						&  4000 		& 2000 (50\%) \\
w1a		& 				& 2477			&  72 (3\%) 						& 47272 		& 1407 (3\%) \\
w2a		& 				& 3470			&  107 (3\%) 						& 46279 		& 1372 (3\%) \\
w3a		& 				& 4912			&  143 (3\%) 						& 44837 		& 1336 (3\%) \\
w4a		& 	300			& 7366			&  216 (3\%) 						& 42383 		& 1263 (3\%) \\
w5a		& 				& 9888			&  281 (3\%) 						& 39861 		& 1198 (3\%) \\
\hline	
\end{tabular}
\caption{LIBSVM UCI data extracts and their characteristics.}
\label{tab:datasets}
\end{table*}

Table \ref{tab:datasets} shows the data characteristics of the 13 data sets used. The fraction of the minority class, $\beta$ ranges from $0.03$ to $0.48$ in training sets and from $0.03$ to $0.50$ in test sets. Also, the feature dimension varies between 4 to 300. Note that although these data sets are not linearly separable, they are still referred to as `clean' before they are subjected to label-manipulation. For generating different types of attacks, $\rho=0.75$ was used while the value of $\alpha$ was varied between $[0.0,~1.0]$ in steps of $0.25$.
We compare \subsvms\ against of four other SVM-based classifiers:
\begin{enumerate}

\item \texttt{Oracle-SVM:} Standard SVM learnt over training data with parameters fixed by cross-validating directly over \textit{clean} test set.

\item \texttt{Blind-SVM:} Standard SVM learnt over training data with parameters fixed based on the best average performance over \textit{all} test sets. This is similar to \texttt{Oracle-SVM}, except that a single set of parameters is used for all data sets. This helps assess the feasibility of blindly fixing the same set of parameters for all test sets.

\item \texttt{Bag-SVM:} Regular bagging of SVMs where each SVM in the ensemble is trained on a bootstrap sample of size same as the original data (sampled with replacement). All SVMs use the same set of optimum parameters, which were determined through test set cross-validation of \texttt{Oracle-SVM}.

\item \texttt{CV-SVM:} Standard SVM with parameters chosen through four-fold cross-validation on the training data. In all the experiments, the results of \texttt{CV-SVM} are averaged over five different random splits of the training data for cross-validation.

\end{enumerate}
All cross-validations were performed by varying the penalty parameter $C$ between 1 and 100, ratio of the weights of the two classes $W$ between 0.1 and 10 and the RBF kernel parameter $\sigma^2$ between $0.1/d$ and $10/d$, where $d$ is the data dimensionality. For \subsvms, the values of $C=100$, $w=1$, $\sigma^2 = 1/d$, $s=\log^2 \ell$ and $J = 1000$ were fixed for all data sets without performing any sort of cross-validation. All the SVMs were trained under L-2 loss, although similar results were also obtained under L-1 loss. 

Note that \texttt{Oracle-SVM}, \texttt{Blind-SVM} and \texttt{Bag-SVM} use information about test set labels to obtain their corresponding set of optimum parameters for training. This gives them an unfair advantage over \texttt{CV-SVM} and \subsvms\ that are both agnostic to test set labels. 

\textbf{Performance measure:}
The UCI data sets exhibit a wide range of class imbalance - a1a--a5a are moderately imbalanced, splice, mushrooms and svmguide1 are class-balanced while w1a--w5a are highly imbalanced. For imbalanced data, high accuracies can be trivially achieved by labeling all points with the majority class label. Since accuracy is ineffective in such settings, we use its skew-insensitive version called Balanced Accuracy\footnote{See Appendix~\ref{sec:performance-metrics} for details of this measure.} (BAC) \cite{brodersen10}. Note that for class-balanced data, BAC reduces to accuracy.

\begin{table*}[!t]
\centering
\begin{tabular}{|@{\hspace{0.02cm}}c@{\hspace{0.02cm}}|@{\hspace{0.05cm}}l@{\hspace{0.05cm}}|@{\hspace{0.05cm}}c@{\hspace{0.05cm}}|@{\hspace{0.05cm}}c@{\hspace{0.05cm}}|@{\hspace{0.05cm}}c@{\hspace{0.05cm}}|@{\hspace{0.05cm}}c@{\hspace{0.05cm}}|@{\hspace{0.05cm}}c@{\hspace{0.05cm}}|@{\hspace{0.05cm}}c@{\hspace{0.05cm}}|@{\hspace{0.05cm}}c@{\hspace{0.05cm}}|@{\hspace{0.05cm}}c@{\hspace{0.05cm}}|@{\hspace{0.05cm}}c@{\hspace{0.05cm}}|@{\hspace{0.05cm}}c@{\hspace{0.05cm}}|@{\hspace{0.05cm}}c@{\hspace{0.05cm}}|@{\hspace{0.05cm}}c@{\hspace{0.05cm}}|@{\hspace{0.05cm}}c@{\hspace{0.05cm}}|}
\toprule
& \textit{Method} & a1a & a2a & a3a & a4a & a5a & splc & mush & svm1 & w1a  & w2a & w3a & w4a & w5a\\
\midrule
\multirow{5}{*}{\textit{clean}} & \texttt{Oracle-SVM} & 0.76 & 0.77 & 0.76 & 0.76 & 0.76 & 0.91 & 1.00 & 0.97 & 0.75 & 0.76 & 0.78 & 0.79 & 0.81\\
& \texttt{Blind-SVM} & 0.75 & 0.75 & 0.75 & 0.76 & 0.76 & 0.89 & 1.00 & 0.97 & 0.75 & 0.76 & 0.78 & 0.79 & 0.81\\
& \texttt{Bag-SVM} & 0.76 & 0.77 & 0.76 & 0.76 & 0.76 & 0.90 & 1.00 & 0.97 & 0.73 & 0.75 & 0.77 & 0.79 & 0.79\\
& \texttt{CV-SVM} & 0.76 & 0.76 & 0.76 & 0.76 & 0.76 & 0.90 & 1.00 & 0.97 & 0.70 & 0.69 & 0.71 & 0.72 & 0.73\\
& \subsvms & \textbf{0.81} & 0.81 & \textbf{0.81} & \textbf{0.81} & \textbf{0.81} & 0.86 & 0.98 & 0.96 & \textbf{0.85} & \textbf{0.86} & \textbf{0.86} & \textbf{0.87} & \textbf{0.88}\\
\midrule
\multirow{5}{*}{$\alpha=1.0$} & \texttt{Oracle-SVM} & 0.79 & 0.80 & 0.80 & 0.81 & 0.81 & 0.57 & 0.64 & 0.89 & 0.74 & 0.77 & 0.78 & 0.79 & 0.81\\
& \texttt{Blind-SVM} & 0.74 & 0.74 & 0.75 & 0.77 & 0.77 & 0.57 & 0.54 & 0.84 & 0.74 & 0.76 & 0.77 & 0.79 & 0.80\\
& \texttt{Bag-SVM} & 0.79 & 0.80 & 0.80 & 0.81 & 0.81 & 0.55 & 0.62 & 0.89 & 0.73 & 0.76 & 0.78 & 0.79 & 0.80\\
& \texttt{CV-SVM} & 0.78 & 0.78 & 0.79 & 0.79 & 0.79 & 0.54 & 0.54 & 0.89 & 0.67 & 0.69 & 0.69 & 0.73 & 0.74\\
& \subsvms & 0.79 & 0.79 & 0.79 & 0.80 & 0.80 & \textbf{0.77} & \textbf{0.97} & 0.90 & \textbf{0.79} & \textbf{0.82} & \textbf{0.83} & \textbf{0.85} & 0.85\\
\midrule
\multirow{5}{*}{$\alpha=0.5$} & \texttt{Oracle-SVM} & 0.63 & 0.63 & 0.63 & 0.64 & 0.64 & 0.73 & 0.99 & 0.90 & 0.64 & 0.65 & 0.67 & 0.68 & 0.69\\
& \texttt{Blind-SVM} & 0.63 & 0.63 & 0.63 & 0.64 & 0.64 & 0.62 & 0.85 & 0.88 & 0.64 & 0.64 & 0.65 & 0.66 & 0.66\\
& \texttt{Bag-SVM} & 0.64 & 0.64 & 0.64 & 0.64 & 0.64 & 0.73 & 0.99 & 0.90 & 0.63 & 0.64 & 0.66 & 0.67 & 0.68\\
& \texttt{CV-SVM} & 0.63 & 0.63 & 0.63 & 0.64 & 0.64 & 0.70 & 0.94 & 0.90 & 0.59 & 0.56 & 0.58 & 0.59 & 0.58\\
& \subsvms & \textbf{0.79} & \textbf{0.79} & \textbf{0.79} & \textbf{0.80} & \textbf{0.80} & 0.74 & 0.98 & 0.93 & \textbf{0.78} & \textbf{0.80} & \textbf{0.81} & \textbf{0.84} & \textbf{0.85}\\
\midrule
\multirow{5}{*}{$\alpha=0.0$} & \texttt{Oracle-SVM} & 0.55 & 0.55 & 0.55 & 0.55 & 0.55 & 0.55 & 0.62 & 0.50 & 0.54 & 0.55 & 0.55 & 0.55 & 0.55\\
& \texttt{Blind-SVM} & 0.55 & 0.55 & 0.55 & 0.55 & 0.55 & 0.53 & 0.62 & 0.50 & 0.54 & 0.55 & 0.55 & 0.55 & 0.55\\
& \texttt{Bag-SVM} & 0.53 & 0.53 & 0.54 & 0.54 & 0.54 & 0.54 & 0.59 & 0.50 & 0.53 & 0.54 & 0.54 & 0.54 & 0.54\\
& \texttt{CV-SVM} & 0.50 & 0.50 & 0.50 & 0.50 & 0.50 & 0.50 & 0.50 & 0.50 & 0.53 & 0.52 & 0.52 & 0.51 & 0.51\\
& \subsvms & \textbf{0.78} & \textbf{0.79} & \textbf{0.80} & \textbf{0.80} & \textbf{0.80} & \textbf{0.79} & \textbf{0.98} & \textbf{0.96} & \textbf{0.74} & \textbf{0.78} & \textbf{0.81} & \textbf{0.82} & \textbf{0.83}\\
\bottomrule
\end{tabular}
\caption{Balanced Accuracy (BAC) under L-2 loss for clean and noisy versions of UCI data sets. For different types of attacks (different $\alpha$) the results for each method are averaged over 10 different noisy versions. `splc', `mush' and `svm1' stand for splice, mushrooms and svmguide1. Only \texttt{CV-SVM} and \subsvms\ are agnostic to the true test labels. The cases where one of the methods is significantly better than all others $(\geq 0.05)$ are highlighted.}
\label{table:L2resultsBAC}
\end{table*}

\begin{table*}[!t]
\centering
\begin{tabular}{|@{\hspace{0.1cm}}l@{\hspace{0.1cm}}|@{\hspace{0.1cm}}r@{\hspace{0.1cm}}|@{\hspace{0.1cm}}r@{\hspace{0.1cm}}|@{\hspace{0.1cm}}r@{\hspace{0.1cm}}|@{\hspace{0.1cm}}r@{\hspace{0.1cm}}|@{\hspace{0.1cm}}r@{\hspace{0.1cm}}|@{\hspace{0.1cm}}r@{\hspace{0.1cm}}|@{\hspace{0.1cm}}r@{\hspace{0.1cm}}|@{\hspace{0.1cm}}r@{\hspace{0.1cm}}|@{\hspace{0.1cm}}r@{\hspace{0.1cm}}|@{\hspace{0.1cm}}r@{\hspace{0.1cm}}|@{\hspace{0.1cm}}r@{\hspace{0.1cm}}|@{\hspace{0.1cm}}r@{\hspace{0.1cm}}|@{\hspace{0.1cm}}r@{\hspace{0.1cm}}|}
\toprule
\textit{Method} & a1a & a2a & a3a & a4a & a5a & splc & mush & svm1 & w1a  & w2a & w3a & w4a & w5a\\
\midrule
\texttt{Oracle-SVM} & 218 & 318 & 456 & 721 & 1432 & 43 & 561 & 44 & 256 & 374 & 522 & 886 & 1433\\
\texttt{Blind-SVM} &  216 & 314 & 450 & 718 & 1420 & 43 & 558 & 41 & 254 & 368 & 514 & 869 & 1416\\
\texttt{Bag-SVM} & 65 & 145 & 293 & 594 & 2287 & 63 & 1546 & 208 & 35 & 72 & 132 & 351 & 839\\
\texttt{CV-SVM} & 170 & 351 & 715 & 1804 & 3501 & 202 & 3276 & 363 & 95 & 196 & 384 & 915 & 1688\\
\subsvms & 3 & 4 & 4 & 5 & 6 & 5 & 8 & 4 & 4 & 5 & 4 & 5 & 6\\
\bottomrule
\end{tabular}
\caption{Training times in seconds (rounded to the closest integer) for all the methods trained using L-2 loss averaged over different corruptions corresponding to the results presented in Table \ref{table:L2resultsBAC}. Note that for \subsvms, the reported time is the time taken to train all the $J=1000$ SVMs on $s=\log^2\ell$-size subsets.}
\label{table:L2RunTimesBAC}
\end{table*}

Table \ref{table:L2resultsBAC} summarizes the results of all the five methods on clean as well as corrupted versions of the data. For every data set, 10 random corruptions were performed w.r.t. the corresponding attack direction $\alpha$ and the averaged results are reported. Winning results, when significantly better than the rest, are highlighted\footnote{Std. devs. were negligible (mostly $<0.02$, max 0.06).}. 

\begin{itemize}

\item \subsvms\ is almost always significantly better than all the other methods (by $5\%$ or more) and is never significantly worse. 
The advantage of \subsvms\ is visible in both balanced and imbalanced data; for imbalanced data, the advantage increases for smaller $\alpha$. This is because the quality of minority-class data falls sharply with $\alpha$.

\item \texttt{Oracle-SVM} is at least as good as \texttt{Blind-SVM}. This is because \texttt{Oracle-SVM} tunes parameters individually for each test set, while \texttt{Blind-SVM} fixes the same parameters across all test sets.

\item \texttt{Oracle-SVM}, \texttt{Blind-SVM} and \texttt{Bag-SVM} are better than \texttt{CV-SVM}. This is because all three methods cross-validate directly on the test sets.

\item \texttt{Bag-SVM}'s performance is similar to that of \texttt{Oracle-SVM}. This is consistent with \cite{valentini04} that also reported no benefit in bagging SVMs (since SVMs are stable classifiers).

\item \texttt{CV-SVM} is the worst performing method and is often significantly worse than others\footnote{The case of clean, balanced data is the only exception.}. This shows its ineffectiveness under noisy settings.

\end{itemize}


Similar results were also obtained using Skew-Insensitive F-score (SIF) \cite{flach03}. Results using Area Under the Curve (AUC) and accuracy, their unsuitability for imbalanced data notwithstanding, are reported in Appendix~\ref{sec:additional-results-L2Loss}.

\textbf{Run-times:} Table \ref{table:L2RunTimesBAC} summarizes training times averaged over different types of attacks. \subsvms~ is clearly much faster than all other methods\footnote{See Appendix~\ref{sec:additional-results-L2Loss} for more detailed run-times.}. While our experiments were based on single-core implementations, \subsvms\ can be easily parallelized to handle very large-scale problems.

\section{Conclusions}
\label{sec:conclusions}
We present a simple algorithm (\subsvms) for learning binary classifiers under adversarial label-noise.
\subsvms\ can efficiently correct a bounded number of adversarial label-errors introduced in linearly separable data.
Extensions to handle attribute noise and multi-class settings are important directions for future work. It would also be interesting to explore applicability of \subsvms\ for solving large, noisy, real-world problems, where SVMs typically perform poorly.

\newpage

\appendix

\section{Error-rate of \texorpdfstring{$\Psi_{\widehat{\SB}}$}{lg} w.r.t. samples drawn uniformly from \texorpdfstring{$\widehat{\DB}$}{lg}}
\label{sec:proof-error-rate}

Let $\epsilon_1$ and $\epsilon_2$ be the class conditional error rates for the two classes. Without loss of generality let $\epsilon_2 \geq \epsilon_1$. In the absence of the knowledge whether $\epsilon_2$ is associated with the minority class or the majority class, the overall error rate of $\Psi_{\widehat{\SB}}$ w.r.t. samples drawn {\em iid} from $\scrD_{p\widehat{\DB}}$ is given by
\begin{equation}
\epsilon = \max\left(p\epsilon_1 + (1-p)\epsilon_2, (1-p)\epsilon_1 + p\epsilon_2\right) \leq \epsilon_2.
\end{equation}
Therefore, if $\epsilon = p\epsilon_1 + (1-p)\epsilon_2$, then
\begin{equation}
\epsilon_2 = \frac{\epsilon-p\epsilon_1}{1-p} \leq \frac{\epsilon}{1-p}
\end{equation}
and if $\epsilon = (1-p)\epsilon_1 + p\epsilon_2$, then
\begin{equation}
\epsilon_2 = \frac{\epsilon-(1-p)\epsilon_1}{p} \leq \frac{\epsilon}{p}.
\end{equation}
Therefore, 
\begin{equation}
\epsilon_2 < \max\left(\frac{\epsilon}{1-p}, \frac{\epsilon}{p}\right) = \frac{\epsilon}{p^\ast}
\end{equation}
where $p^\ast=\min\{p,1-p\}$.


\section{Proof for optimality of class-balanced sampling (\texorpdfstring{$p=0.5$}{lg})}
\label{sec:proof-optimality-cb}

Consider a two-class classification problem where the two classes are represented by $A$ and $B$. Without loss of generality, let $A$ be the minority class containing $0<\beta\leq0.5$  fraction of the points. Let $\widehat{A}$ and $\widehat{B}$ represent the two classes after one or both the classes are corrupted with adversarial noise. Let $\rho\beta,~0\leq\rho<1$ represent the upper limit on the fraction of corrupted points. Therefore, the total number of
corrupted points can be written as $n_c = \rho\beta\ell$. Further, let $\alpha$ be the fraction of the corrupted points that were originally in class $B$ but were assigned to class $A$. Therefore, the fraction of the new classes can be given by
\begin{align}
|\widehat{A}| &= \beta + \alpha\rho\beta - (1-\alpha)\rho\beta\\
|\widehat{B}| &= 1 - \beta - \alpha\rho\beta + (1-\alpha)\rho\beta
\end{align}
Moreover, the fraction of good (clean) and bad (mislabeled) points in both the classes are
\begin{align}
|\widehat{A}_g| &= \beta - (1-\alpha)\rho\beta\\
|\widehat{A}_b| &= \alpha\rho\beta\\
|\widehat{B}_g| &= 1 - \beta - \alpha\rho\beta\\
|\widehat{B}_b| &= (1-\alpha)\rho\beta
\end{align}
Therefore, the conditional probability of picking a good or a bad point for both the classes are given by
\begin{align}
P(a_g|\widehat{A}) &= \frac{|\widehat{A}_g|}{|\widehat{A}|} = \frac{1 - (1-\alpha)\rho}{1 + \alpha\rho - (1-\alpha)\rho}\\
P(a_b|\widehat{A}) &= \frac{|\widehat{A}_b|}{|\widehat{A}|} = \frac{\alpha\rho}{1 + \alpha\rho - (1-\alpha)\rho}\\
P(b_g|\widehat{B}) &= \frac{|\widehat{B}_g|}{|\widehat{B}|} = \frac{1 - \beta - \alpha\rho\beta}{1 - \beta - \alpha\rho\beta + (1-\alpha)\rho\beta}\\
P(b_b|\widehat{B}) &= \frac{|\widehat{B}_b|}{|\widehat{B}|} = \frac{(1-\alpha)\rho\beta}{1 - \beta - \alpha\rho\beta + (1-\alpha)\rho\beta}
\end{align}
Assuming that the probability with which points from classes $\widehat{A}$ and $\widehat{B}$ are picked is given by $P(\widehat{A}) = p$ and $P(\widehat{B}) = 1-p$ respectively, the probability of picking up a good or a bad point for both the classes are respectively given by
$P(a_g) = P(a_g|\widehat{A})p$, $P(a_b) = P(a_b|\widehat{A})p$, $P(b_g) = P(b_g|\widehat{B})(1-p)$ and $P(b_b) = P(b_b|\widehat{B})(1-p)$.

The probability $\eta$ of not picking $r/2$ clean points from either class is upper bounded by
\begin{equation}
\eta \leq \sum_{k=0}^{r/2-1} {s \choose k} \left(P(a_g)\right)^k \left(1 - P(a_g)\right)^{s-k} + \sum_{k=0}^{r/2-1} {s \choose k} \left(P(b_g)\right)^k \left(1 - P(b_g)\right)^{s-k}.
\label{eqn:etaBinomial1}
\end{equation}
For worst case analysis, we need to maximize $\eta$ and therefore, minimize both $P(a_g)$ and $P(b_g)$, which in turn requires minimizing $P(a_g|\widehat{A})$ and $P(b_g|\widehat{B})$ w.r.t. both $\alpha$ and $\beta$. Differentiating $P(a_g|\widehat{A})$ w.r.t $\alpha$
\begin{equation}
\frac{dP(a_g|\widehat{A})}{d\alpha} = \frac{\rho(-1 + \rho)}{(1 - \rho + 2\alpha\rho)^2} \leq 0.
\end{equation}
Therefore, 
\begin{equation}
\arg\min_\alpha P(a_g) = 1.
\end{equation}
Similarly, differentiating $P(b_g|\widehat{B})$ w.r.t. $\alpha$
\begin{equation}
\frac{dP(b_g|\widehat{B})}{d\alpha} = \frac{\rho\beta(1 - \beta(1+\rho))}{(1 - \beta + \rho\beta - 2\alpha\rho\beta)^2} \geq 0
\end{equation}
Therefore, 
\begin{equation}
\arg\min_\alpha P(b_g) = 0.
\end{equation}
Also,
\begin{equation}
\frac{dP(b_g|\widehat{B})}{d\alpha} = \frac{-\rho(1 - \alpha)}{(1 - \beta + \rho\beta - 2\alpha\rho\beta)^2} \leq 0
\end{equation}
implying that
\begin{equation}
\arg\min_\beta P(b_g) = \frac{1}{2} ~.
\end{equation}
Substituting $\alpha = 1$ in $P(a_g)$ and $\alpha = 0, \beta = 1/2$ in $P(b_g)$, we get
\begin{equation}
\min P(a_g) = \frac{p}{1+\rho} ~~~~\text{and}~~~~~ \min P(b_g) = \frac{1-p}{1+\rho}.
\end{equation}
Therefore, the worst case bound for (\ref{eqn:etaBinomial1}) can be written as
\begin{equation}
\label{eqn:etaBinomial2}
\eta \leq \sum_{k=0}^{r/2-1} {s \choose k} \left(\frac{p}{1+\rho}\right)^k \left(1 - \frac{p}{1+\rho}\right)^{s-k} + \sum_{k=0}^{r/2-1} {s \choose k} \left(\frac{1-p}{1+\rho}\right)^k \left(1 - \frac{1-p}{1+\rho}\right)^{s-k}.
\end{equation}

Applying Hoeffding bound \cite{hoeffding63} individually on each of the two terms
\begin{equation}
\label{eqn:etaHoeffding1}
\eta \leq \frac{1}{2} \exp{\left(-\frac{2}{s}\left(\frac{sp}{1-\rho} - \frac{r}{2} + 1\right)^2\right)} + \frac{1}{2} \exp{\left(-\frac{2}{s}\left(\frac{s(1-p)}{1+\rho} - \frac{r}{2} + 1\right)^2\right)}
\end{equation}
as long as $\frac{sp}{1+\rho} > \frac{r}{2} - 1$ and $\frac{s(1-p)}{1+\rho} > \frac{r}{2} - 1$. The RHS of \ref{eqn:etaHoeffding1} can be rewritten as 
\begin{equation}
\label{eqn:etaHoeffding2}
f = \frac{1}{2}\exp\left(-\frac{1}{2}\left(\frac{p - \frac{(r-2)(1+\rho)}{2s}}{\left(\frac{1+\rho}{2\sqrt{s}}\right)}\right)^2\right) + \frac{1}{2}\exp\left(-\frac{1}{2}\left(\frac{(1-p) - \frac{(r-2)(1+\rho)}{2s}}{\left(\frac{1+\rho}{2\sqrt{s}}\right)}\right)^2\right)
\end{equation}
which is simply the sum of two Gaussian with means $\mu_1 = \frac{(r-2)(1+\rho)}{2s}$ and $\mu_2 = 1-\frac{(r-2)(1+\rho)}{2s}$ and equal variance $\sigma = \frac{1+\rho}{2\sqrt{s}}$. Differentiating the above expression w.r.t. $p$
\begin{equation}
\label{eqn:df}
\frac{df}{dp} = - \frac{2\left(\frac{sp}{1 + \rho} - \frac{r}{2} + 1\right)}{\exp\left(\frac{2}{s}\left(\frac{sp}{1 + \rho} - \frac{r}{2} + 1\right)^2\right)(1 + \rho)} + \frac{2\left(\frac{s(1 - p)}{1 + \rho} -\frac{r}{2} + 1\right)}{\exp\left(\frac{2}{s}\left(\frac{s(1 - p)}{1 + \rho} - \frac{r}{2} + 1\right)^2\right)(1 + \rho)}.
\end{equation}
It can be clearly seen that $p=0.5$ is a solution of (\ref{eqn:df}). Also, the sum of two Gaussians can be either unimodal ($p=0.5$ is global maximum) or bimodal ($p=0.5$ is a minimum) \cite{behboodian70}. The second order derivative of $f$ w.r.t. $p$ can be written as
\begin{equation}
\frac{d^2f}{dp} = \frac{8\left(\frac{sp}{1 + \rho} - \frac{r}{2} + 1\right)^2 - 2s}{\exp\left(\frac{2}{s}\left(\frac{sp}{1 + \rho} - \frac{r}{2} + 1\right)^2\right)(1 + \rho)^2}
+ \frac{8\left(\frac{s(1 - p)}{1 + \rho} - \frac{r}{2} + 1\right)^2 - 2s}{\exp\left(\frac{2}{s}\left(\frac{s(1 - p)}{1 + \rho} - \frac{r}{2} + 1\right)^2\right)(1 + \rho)^2}.
\end{equation}
Therefore, enforcing a minimum at $p=0.5$, we get the condition that
\begin{equation}
\frac{d^2f}{dp}\Big |_{p=0.5} = \frac{32\left(\frac{s}{2(1 + \rho)} - \frac{r}{2} + 1\right)^2}{\exp\left(\frac{2}{s}\left(\frac{s}{2(1 + \rho)} - \frac{r}{2} + 1\right)^2\right)(1 + \rho)^2} - \frac{8s}{\exp\left(\frac{2}{s}\left(\frac{s}{2(1 + \rho)} - \frac{r}{2} + 1\right)^2\right)(1 + \rho)^2} \geq 0.
\end{equation}
This directly implies that
\begin{equation}
s \geq (\rho + 1)\left(r -2 + \frac{1}{2}(\rho + 1)\left(1 + \left(\frac{4r + \rho - 7}{\rho + 1}\right)^{1/2}\right)\right)
\label{eqn:s-final}
\end{equation}
which, as expected, is a stronger condition than the one required for imposing the Hoeffding bound at $p = 0.5$, i.e., $s > (1+\rho)(r - 2)$. Furthermore, to enforce $p=0.5$ to be the \textit{global} minimum, we impose the condition that the value of $f$ at $p=0.5$ is strictly less than that at any of the two extreme points of $f$ (i.e., at $p = \frac{1}{s}(1+\rho)(r/2-1)$ and $p = 1 - \frac{1}{s}(1+\rho)(r/2-1)$). This gives us an even stronger condition
\begin{equation}
s \geq (\rho + 1)\left(r -2 + (\rho + 1)\left(\log2 + \left(\frac{\log2(\log2 + 2r + \rho\log2 - 4)}{\rho + 1}\right)^{1/2}\right)\right).
\end{equation}

This is the sufficient condition to guarantee that the worst case probability of selecting less than $r/2$ clean points per class is minimum at $p=0.5$, i.e., when \textit{class-balanced} sampling is performed over the data.

%
%

\section{Details of performance metrics}
\label{sec:performance-metrics}
For class-imbalanced data sets, very high classification accuracy can be trivially obtained by labeling the entire data with the majority class label. The use of Balanced Accuracy (BAC) for class-imbalanced data sets is prescribed by \cite{brodersen10} and can be simply computed as
\begin{equation}
\text{BAC} = \frac{sensitivity + specificity}{2}.
\end{equation}
The $sensitivity$ and $specificity$ are defined as follows
\begin{align}
sensitivity & = \frac{tp}{tp + fn}\\
specificity & = \frac{tn}{tp + fn}
\end{align} 
where $tp$ and $fp$ denote the number of true and false positives while $tn$ and $fn$ denote the number of true and false negatives.

Similarly, traditional F-score can be trivially maximized for imbalanced data sets by compromising recall for high precision. Therefore, SIF \cite{flach03} serves as an alternative to the F-score for imbalanced data sets and is given by
\begin{equation}
\text{SIF} = \frac{2tpr}{tpr + fpr + 1}
\end{equation}
where $tpr$ and $fpr$ are true and false positive rates respectively. Like BAC, SIF also reduces to traditional F-score for class-balanced data sets. Another popular metric for comparison of classification performances is Area Under the ROC Curve (AUC). Although, unlike BAC and SIF, AUC is not a skew-insensitive measure, we also computed AUC measures for all the methods. It is important to mention that \subsvms is always comparable to that of the other methods w.r.t. AUC. Finally, we note that for the results reported using AUC, we needed to retrain an SVM on the error-corrected data (unlike earlier, when we directly used majority voting on the test data).

\section{Additional Results}
\label{sec:additional-results-L2Loss}

Tables \ref{table:L2resultsFSCORE}, \ref{table:L2resultsAUC} and \ref{table:L2resultsACC} present additional results on the UCI data sets under L-2 loss using Skew-Insensitive F-Score (SIF), Area Under the Curve (AUC) and Accuracy, respectively. Table \ref{table:L2DetailedRunTimesBAC} shows detailed run-times corresponding to Table \ref{table:L2RunTimesBAC} in the paper.

\begin{table*}
\centering
\begin{tabular}{|@{\hspace{0.02cm}}c@{\hspace{0.02cm}}|@{\hspace{0.05cm}}l@{\hspace{0.05cm}}|@{\hspace{0.05cm}}c@{\hspace{0.05cm}}|@{\hspace{0.05cm}}c@{\hspace{0.05cm}}|@{\hspace{0.05cm}}c@{\hspace{0.05cm}}|@{\hspace{0.05cm}}c@{\hspace{0.05cm}}|@{\hspace{0.05cm}}c@{\hspace{0.05cm}}|@{\hspace{0.05cm}}c@{\hspace{0.05cm}}|@{\hspace{0.05cm}}c@{\hspace{0.05cm}}|@{\hspace{0.05cm}}c@{\hspace{0.05cm}}|@{\hspace{0.05cm}}c@{\hspace{0.05cm}}|@{\hspace{0.05cm}}c@{\hspace{0.05cm}}|@{\hspace{0.05cm}}c@{\hspace{0.05cm}}|@{\hspace{0.05cm}}c@{\hspace{0.05cm}}|@{\hspace{0.05cm}}c@{\hspace{0.05cm}}|}
\toprule
& \textit{Method} & a1a & a2a & a3a & a4a & a5a & splc & mush & svm1 & w1a  & w2a & w3a & w4a & w5a\\
\midrule
\multirow{5}{*}{\textit{clean}} & \texttt{Oracle-SVM} & 0.71 & 0.73 & 0.71 & 0.72 & 0.72 & 0.91 & 1.00 & 0.97 & 0.66 & 0.69 & 0.72 & 0.74 & 0.76\\
& \texttt{Blind-SVM} & 0.71 & 0.72 & 0.71 & 0.71 & 0.71 & 0.89 & 1.00 & 0.97 & 0.66 & 0.68 & 0.72 & 0.74 & 0.76\\
& \texttt{Bag-SVM} & 0.71 & 0.73 & 0.72 & 0.72 & 0.72 & 0.90 & 1.00 & 0.97 & 0.63 & 0.67 & 0.69 & 0.73 & 0.74\\
& \texttt{CV-SVM} & 0.71 & 0.72 & 0.71 & 0.71 & 0.72 & 0.90 & 1.00 & 0.97 & 0.55 & 0.56 & 0.55 & 0.61 & 0.64\\
& \subsvms & \textbf{0.81} & \textbf{0.82} & \textbf{0.82} & \textbf{0.82} & \textbf{0.82} & 0.87 & 0.98 & 0.96 & \textbf{0.85} & \textbf{0.86} & \textbf{0.86} & \textbf{0.88} & \textbf{0.88}\\
\midrule
\multirow{5}{*}{$\alpha=1.0$} & \texttt{Oracle-SVM} & 0.79 & 0.80 & 0.80 & 0.81 & 0.81 & 0.70 & 0.74 & 0.90 & 0.66 & 0.70 & 0.72 & 0.74 & 0.76\\
& \texttt{Blind-SVM} & 0.74 & 0.75 & 0.75 & 0.77 & 0.77 & 0.70 & 0.68 & 0.87 & 0.66 & 0.69 & 0.70 & 0.74 & 0.75\\
& \texttt{Bag-SVM} & 0.79 & 0.80 & 0.80 & 0.80 & 0.81 & 0.69 & 0.73 & 0.90 & 0.64 & 0.69 & 0.71 & 0.74 & 0.75\\
& \texttt{CV-SVM} & 0.77 & 0.79 & 0.79 & 0.79 & 0.79 & 0.69 & 0.69 & 0.90 & 0.64 & 0.59 & 0.60 & 0.67 & 0.70\\
& \subsvms & 0.81 & 0.81 & 0.81 & 0.82 & 0.82 & \textbf{0.81} & \textbf{0.97} & 0.91 & \textbf{0.79} & \textbf{0.84} & \textbf{0.84} & \textbf{0.85} & \textbf{0.86}\\
\midrule
\multirow{5}{*}{$\alpha=0.5$} & \texttt{Oracle-SVM} & 0.51 & 0.52 & 0.52 & 0.52 & 0.52 & 0.72 & 0.99 & 0.89 & 0.45 & 0.48 & 0.51 & 0.53 & 0.55\\
& \texttt{Blind-SVM} & 0.51 & 0.52 & 0.52 & 0.52 & 0.52 & 0.64 & 0.65 & 0.89 & 0.41 & 0.47 & 0.51 & 0.53 & 0.55\\
& \texttt{Bag-SVM} & 0.50 & 0.51 & 0.51 & 0.51 & 0.51 & 0.72 & 0.99 & 0.89 & 0.41 & 0.43 & 0.48 & 0.50 & 0.53\\
& \texttt{CV-SVM} & 0.25 & 0.21 & 0.31 & 0.34 & 0.34 & 0.71 & 0.96 & 0.71 & 0.32 & 0.23 & 0.29 & 0.31 & 0.29\\
& \subsvms & \textbf{0.80} & \textbf{0.80} & \textbf{0.81} & \textbf{0.81} & \textbf{0.82} & 0.75 & 0.98 & \textbf{0.94} & \textbf{0.78} & \textbf{0.81} & \textbf{0.81} & \textbf{0.84} & \textbf{0.85}\\
\midrule
\multirow{5}{*}{$\alpha=0.0$} & \texttt{Oracle-SVM} & 0.22 & 0.22 & 0.23 & 0.22 & 0.22 & 0.20 & 0.37 & 0.01 & 0.16 & 0.17 & 0.19 & 0.18 & 0.18\\
& \texttt{Blind-SVM} & 0.22 & 0.22 & 0.23 & 0.22 & 0.22 & 0.13 & 0.37 & 0.01 & 0.14 & 0.17 & 0.18 & 0.18 & 0.18\\
& \texttt{Bag-SVM} & 0.14 & 0.15 & 0.15 & 0.16 & 0.16 & 0.15 & 0.31 & 0.01 & 0.12 & 0.14 & 0.15 & 0.14 & 0.15\\
& \texttt{CV-SVM} & 0.00 & 0.00 & 0.00 & 0.00 & 0.00 & 0.00 & 0.00 & 0.00 & 0.08 & 0.12 & 0.07 & 0.07 & 0.03\\
& \subsvms & \textbf{0.78} & \textbf{0.78} & \textbf{0.79} & \textbf{0.79} & \textbf{0.81} & \textbf{0.77} & \textbf{0.97} & \textbf{0.96} & \textbf{0.69} & \textbf{0.75} & \textbf{0.80} & \textbf{0.80} & \textbf{0.81}\\
\bottomrule
\end{tabular}
\caption{Skew-Insensitive F-score (SIF) under L-2 loss for clean and noisy versions of UCI data sets. For different types of attacks (different $\alpha$) the results for each method are averaged over 10 different noisy versions. `splc', `mush' and `svm1' stand for splice, mushrooms and svmguide1. Only \texttt{CV-SVM} and \subsvms\ are agnostic to the true test labels. The cases where one of the methods is significantly better than all others $(\geq 0.05)$ are highlighted.}
\label{table:L2resultsFSCORE}
\end{table*}

\begin{table*}
\centering
\begin{tabular}{|@{\hspace{0.09cm}}c@{\hspace{0.09cm}}|@{\hspace{0.09cm}}l@{\hspace{0.09cm}}|@{\hspace{0.09cm}}c@{\hspace{0.09cm}}|@{\hspace{0.09cm}}c@{\hspace{0.09cm}}|@{\hspace{0.09cm}}c@{\hspace{0.09cm}}|@{\hspace{0.09cm}}c@{\hspace{0.09cm}}|@{\hspace{0.09cm}}c@{\hspace{0.09cm}}|@{\hspace{0.09cm}}c@{\hspace{0.09cm}}|@{\hspace{0.09cm}}c@{\hspace{0.09cm}}|@{\hspace{0.09cm}}c@{\hspace{0.09cm}}|@{\hspace{0.09cm}}c@{\hspace{0.09cm}}|@{\hspace{0.09cm}}c@{\hspace{0.09cm}}|@{\hspace{0.09cm}}c@{\hspace{0.09cm}}|@{\hspace{0.09cm}}c@{\hspace{0.09cm}}|@{\hspace{0.09cm}}c@{\hspace{0.09cm}}|}
\toprule
& \textit{Method} & a1a & a2a & a3a & a4a & a5a & splc & mush & svm1 & w1a  & w2a & w3a & w4a & w5a\\
\midrule
\multirow{5}{*}{\textit{clean}} & \texttt{Oracle-SVM} & 0.89 & 0.90 & 0.90 & 0.90 & 0.90 & 0.96 & 1.00 & 1.00 & 0.93 & 0.95 & 0.96 & 0.96 & 0.96\\
& \texttt{Blind-SVM} & 0.89 & 0.90 & 0.90 & 0.90 & 0.90 & 0.95 & 1.00 & 0.99 & 0.93 & 0.95 & 0.95 & 0.96 & 0.96\\
& \texttt{Bag-SVM} & 0.88 & 0.89 & 0.89 & 0.89 & 0.89 & 0.96 & 1.00 & 1.00 & 0.80 & 0.88 & 0.90 & 0.91 & 0.94\\
& \texttt{CV-SVM} & 0.89 & 0.90 & 0.90 & 0.90 & 0.90 & 0.96 & 1.00 & 1.00 & 0.91 & 0.95 & 0.96 & 0.96 & 0.96\\
& \subsvms\ & 0.89 & 0.89 & 0.89 & 0.90 & 0.90 & 0.93 & 0.98 & 0.99 & 0.90 & 0.92 & 0.92 & 0.93 & 0.94\\
\midrule
\multirow{5}{*}{$\alpha=1.0$} & \texttt{Oracle-SVM} & 0.88 & 0.89 & 0.89 & 0.89 & 0.89 & 0.88 & 1.00 & 0.99 & 0.92 & 0.93 & 0.94 & 0.94 & 0.95\\
& \texttt{Blind-SVM} & 0.88 & 0.89 & 0.89 & 0.89 & 0.89 & 0.85 & 1.00 & 0.99 & 0.91 & 0.93 & 0.94 & 0.94 & 0.94\\
& \texttt{Bag-SVM} & 0.87 & 0.88 & 0.89 & 0.89 & 0.89 & 0.88 & 0.56 & 0.99 & 0.98 & 0.92 & 0.88 & 0.92 & 0.90\\
& \texttt{CV-SVM} & 0.86 & 0.87 & 0.87 & 0.88 & 0.88 & 0.88 & 0.99 & 0.99 & 0.89 & 0.92 & 0.94 & 0.94 & 0.94\\
& \subsvms\ & 0.88 & 0.88 & 0.89 & 0.89 & 0.89 & 0.88 & 0.98 & 0.99 & 0.86 & 0.90 & 0.91 & 0.92 & 0.93\\
\midrule
\multirow{5}{*}{$\alpha=0.5$} & \texttt{Oracle-SVM} & 0.88 & 0.88 & 0.88 & 0.88 & 0.89 & 0.82 & 1.00 & 0.99 & 0.91 & 0.92 & 0.93 & 0.94 & 0.94\\
& \texttt{Blind-SVM} & 0.88 & 0.88 & 0.88 & 0.88 & 0.89 & 0.82 & 1.00 & 0.96 & 0.91 & 0.92 & 0.92 & 0.93 & 0.94\\
& \texttt{Bag-SVM} & 0.82 & 0.84 & 0.85 & 0.86 & 0.86 & 0.85 & 1.00 & 0.99 & 0.98 & 0.97 & 0.89 & 0.89 & 0.86\\
& \texttt{CV-SVM} & 0.84 & 0.84 & 0.86 & 0.86 & 0.88 & 0.78 & 0.97 & 0.99 & 0.89 & 0.91 & 0.92 & 0.93 & 0.93\\
& \subsvms & 0.87 & 0.87 & 0.88 & 0.88 & 0.89 & 0.82 & 0.99 & 0.99 & 0.86 & 0.88 & 0.89 & 0.91 & 0.92\\
\midrule
\multirow{5}{*}{$\alpha=0.0$} & \texttt{Oracle-SVM} & 0.87 & 0.88 & 0.88 & 0.89 & 0.89 & 0.87 & 1.00 & 0.99 & 0.89 & 0.90 & 0.92 & 0.93 & 0.93\\
& \texttt{Blind-SVM} & 0.87 & 0.88 & 0.88 & 0.89 & 0.89 & 0.87 & 1.00 & 0.98 & 0.88 & 0.90 & 0.92 & 0.93 & 0.93\\
& \texttt{Bag-SVM} & 0.50 & 0.50 & 0.50 & 0.50 & 0.50 & 0.53 & 0.08 & 0.94 & 0.98 & 0.98 & 0.98 & 0.98 & 0.98\\
& \texttt{CV-SVM} & 0.86 & 0.86 & 0.88 & 0.88 & 0.89 & 0.86 & 1.00 & 0.99 & 0.87 & 0.88 & 0.91 & 0.92 & 0.92\\
& \subsvms & 0.87 & 0.88 & 0.88 & 0.89 & 0.89 & 0.88 & 0.99 & 0.99 & 0.83 & 0.86 & 0.88 & 0.90 & 0.91\\
\bottomrule
\end{tabular}
\caption{Area Under the Curve (AUC) under L-2 loss for clean and noisy versions of UCI data sets. For different types of attacks (different $\alpha$) the results for each method are averaged over 10 different noisy versions. `splc', `mush' and `svm1' stand for splice, mushrooms and svmguide1. Only \texttt{CV-SVM} and \subsvms\ are agnostic to the true test labels. The cases where one of the methods is significantly better than all others $(\geq 0.05)$ are highlighted.}
\label{table:L2resultsAUC}
\end{table*}

\begin{table*}
\centering
\begin{tabular}{|@{\hspace{0.085cm}}c@{\hspace{0.085cm}}|@{\hspace{0.085cm}}l@{\hspace{0.085cm}}|@{\hspace{0.085cm}}c@{\hspace{0.085cm}}|@{\hspace{0.085cm}}c@{\hspace{0.085cm}}|@{\hspace{0.085cm}}c@{\hspace{0.085cm}}|@{\hspace{0.085cm}}c@{\hspace{0.085cm}}|@{\hspace{0.085cm}}c@{\hspace{0.085cm}}|@{\hspace{0.085cm}}c@{\hspace{0.085cm}}|@{\hspace{0.085cm}}c@{\hspace{0.085cm}}|@{\hspace{0.085cm}}c@{\hspace{0.085cm}}|@{\hspace{0.085cm}}c@{\hspace{0.085cm}}|@{\hspace{0.085cm}}c@{\hspace{0.085cm}}|@{\hspace{0.085cm}}c@{\hspace{0.085cm}}|@{\hspace{0.085cm}}c@{\hspace{0.085cm}}|@{\hspace{0.085cm}}c@{\hspace{0.085cm}}|}
\toprule
& \textit{Method} & a1a & a2a & a3a & a4a & a5a & splc & mush & svm1 & w1a  & w2a & w3a & w4a & w5a\\
\midrule
\multirow{5}{*}{\textit{clean}} & \texttt{Oracle-SVM} & 0.84 & 0.85 & 0.85 & 0.85 & 0.85 & 0.91 & 1.00 & 0.97 & 0.98 & 0.98 & 0.98 & 0.98 & 0.99\\
& \texttt{Blind-SVM} & 0.84 & 0.84 & 0.84 & 0.85 & 0.85 & 0.90 & 1.00 & 0.96 & 0.98 & 0.98 & 0.98 & 0.98 & 0.98\\
& \texttt{Bag-SVM} & 0.84 & 0.85 & 0.85 & 0.85 & 0.85 & 0.90 & 1.00 & 0.97 & 0.98 & 0.98 & 0.98 & 0.98 & 0.99\\
& \texttt{CV-SVM} & {0.84} & {0.84} & {0.85} & {0.85} & {0.85} & {0.91} & 1.00 & 0.97 & {0.98} & {0.98} & {0.98} & {0.98} & {0.99}\\
& \subsvms & 0.78 & 0.78 & 0.78 & 0.79 & 0.79 & 0.86 & 0.98 & 0.96 & 0.84 & 0.84 & 0.83 & 0.85 & 0.86\\
\midrule
\multirow{5}{*}{$\alpha=1.0$} & \texttt{Oracle-SVM} & 0.80 & 0.79 & 0.81 & 0.81 & 0.81 & 0.56 & 0.63 & 0.89 & 0.98 & 0.98 & 0.98 & 0.98 & 0.99\\
& \texttt{Blind-SVM} & 0.79 & 0.78 & 0.80 & 0.81 & 0.81 & 0.51 & 0.49 & 0.82 & 0.98 & 0.98 & 0.98 & 0.98 & 0.98\\
& \texttt{Bag-SVM} & 0.80 & 0.79 & 0.81 & 0.81 & 0.81 & 0.53 & 0.61 & 0.89 & 0.98 & 0.98 & 0.98 & 0.98 & 0.99\\
& \texttt{CV-SVM} & 0.77 & 0.77 & 0.78 & {0.80} & {0.79} & 0.48 & 0.49 & 0.89 & {0.98} & {0.97} & {0.98} & {0.98} & {0.98}\\
& \subsvms & 0.73 & 0.73 & 0.74 & 0.74 & 0.74 & \textbf{0.77} & \textbf{0.97} & 0.90 & 0.78 & 0.76 & 0.77 & 0.82 & 0.81\\
\midrule
\multirow{5}{*}{$\alpha=0.5$} & \texttt{Oracle-SVM} & 0.80 & 0.80 & 0.81 & 0.81 & 0.81 & 0.74 & 0.99 & 0.90 & 0.98 & 0.98 & 0.98 & 0.98 & 0.98\\
& \texttt{Blind-SVM} & 0.80 & 0.80 & 0.81 & 0.81 & 0.81 & 0.70 & 0.95 & 0.85 & 0.97 & 0.97 & 0.97 & 0.97 & 0.97\\
& \texttt{Bag-SVM} & 0.80 & 0.80 & 0.81 & 0.81 & 0.81 & 0.74 & 0.99 & 0.90 & 0.98 & 0.98 & 0.98 & 0.98 & 0.98\\
& \texttt{CV-SVM} & 0.79 & {0.80} & {0.80} & 0.80 & 0.80 & 0.71 & 0.95 & 0.90 & {0.97} & {0.97} & {0.98} & {0.98} & {0.98}\\
& \subsvms & 0.75 & 0.75 & 0.75 & 0.76 & 0.76 & 0.74 & 0.97 & 0.93 & 0.75 & 0.75 & 0.81 & 0.82 & 0.83\\
\midrule
\multirow{5}{*}{$\alpha=0.0$} & \texttt{Oracle-SVM} & 0.77 & 0.77 & 0.77 & 0.77 & 0.77 & 0.57 & 0.63 & 0.50 & 0.97 & 0.97 & 0.97 & 0.97 & 0.97\\
& \texttt{Blind-SVM} & 0.77 & 0.77 & 0.77 & 0.77 & 0.77 & 0.55 & 0.63 & 0.50 & 0.97 & 0.97 & 0.97 & 0.97 & 0.97\\
& \texttt{Bag-SVM} & 0.77 & 0.77 & 0.77 & 0.77 & 0.77 & 0.56 & 0.61 & 0.50 & 0.97 & 0.97 & 0.97 & 0.97 & 0.97\\
& \texttt{CV-SVM} & 0.76 & 0.76 & 0.76 & 0.76 & 0.76 & 0.52 & 0.52 & 0.50 & {0.97} & {0.97} & {0.97} & {0.97} & 0.97\\
& \subsvms & 0.80 & 0.79 & 0.79 & 0.80 & 0.80 & \textbf{0.79} & \textbf{0.97} & \textbf{0.96} & 0.86 & 0.84 & 0.82 & 0.87 & 0.93\\
\bottomrule
\end{tabular}
\caption{Accuracy under L-2 loss for clean and noisy versions of UCI data sets. For different types of attacks (different $\alpha$) the results for each method are averaged over 10 different noisy versions. `splc', `mush' and `svm1' stand for splice, mushrooms and svmguide1. Only \texttt{CV-SVM} and \subsvms\ are agnostic to the true test labels. The cases where one of the methods is significantly better than all others $(\geq 0.05)$ are highlighted.}
\label{table:L2resultsACC}
\end{table*}

\begin{table*}
\centering
\begin{tabular}{|@{\hspace{0.05cm}}c@{\hspace{0.05cm}}|@{\hspace{0.05cm}}l@{\hspace{0.05cm}}|@{\hspace{0.1cm}}c@{\hspace{0.1cm}}|@{\hspace{0.1cm}}c@{\hspace{0.1cm}}|@{\hspace{0.1cm}}c@{\hspace{0.1cm}}|@{\hspace{0.1cm}}c@{\hspace{0.1cm}}|@{\hspace{0.1cm}}c@{\hspace{0.1cm}}|@{\hspace{0.1cm}}c@{\hspace{0.1cm}}|@{\hspace{0.05cm}}c@{\hspace{0.01cm}}|@{\hspace{0.05cm}}c@{\hspace{0.05cm}}|@{\hspace{0.1cm}}c@{\hspace{0.1cm}}|@{\hspace{0.1cm}}c@{\hspace{0.1cm}}|@{\hspace{0.1cm}}c@{\hspace{0.1cm}}|@{\hspace{0.1cm}}c@{\hspace{0.1cm}}|@{\hspace{0.1cm}}c@{\hspace{0.1cm}}|}
\toprule
& \textit{Method} & a1a & a2a & a3a & a4a & a5a & splc & mush & svm1 & w1a  & w2a & w3a & w4a & w5a\\
\midrule
\multirow{5}{*}{\textit{clean}} & \texttt{Oracle-SVM} & 180 & 266 & 376 & 601 & 937 & 38 & 102 & 19 & 213 & 286 & 372 & 532 & 805\\
& \texttt{Blind-SVM} & 175 & 258 & 360 & 588 & 917 & 38 & 81 & 17 & 207 & 269 & 347 & 483 & 746\\
& \texttt{Bag-SVM} & 64 & 134 & 276 & 617 & 1675 & 64 & 167 & 38 & 26 & 51 & 103 & 215 & 369\\
& \texttt{CV-SVM} & 138 & 286 & 579 & 1465 & 2860 & 190 & 1033 & 102 & 82 & 160 & 293 & 654 & 1146\\
& \subsvms & 3 & 3 & 3 & 4 & 4 & 4 & 7 & 3 & 3 & 3 & 3 & 4 & 5\\
\midrule
\multirow{5}{*}{$\alpha=1.0$} & \texttt{Oracle-SVM} & 291 & 418 & 606 & 957 & 2211 & 47 & 372 & 61 & 387 & 568 & 840 & 1521 & 2528\\
& \texttt{Blind-SVM} & 292 & 420 & 610 & 960 & 2213 & 47 & 370 & 59 & 383 & 568 & 840 & 1520 & 2527\\
& \texttt{Bag-SVM} & 73 & 147 & 300 & 703 & 3272 & 69 & 1179 & 358 & 59 & 129 & 224 & 710 & 1951\\
& \texttt{CV-SVM} & 242 & 499 & 1034 & 2616 & 5036 & 214 & 3143 & 532 & 157 & 329 & 678 & 1633 & 3029\\
& \subsvms & 4 & 5 & 5 & 6 & 7 & 5 & 8 & 4 & 4 & 4 & 5 & 6 & 7\\
\midrule
\multirow{5}{*}{$\alpha=0.5$} & \texttt{Oracle-SVM} & 281 & 408 & 588 & 931 & 1980 & 55 & 1513 & 67 & 304 & 460 & 639 & 1144 & 1905\\
& \texttt{Blind-SVM} & 280 & 408 & 590 & 938 & 1980 & 55 & 1507 & 67 & 304 & 459 & 640 & 1144 & 1908\\
& \texttt{Bag-SVM} & 100 & 256 & 518 & 860 & 3859 & 72 & 3937 & 330 & 39 & 76 & 146 & 365 & 828\\
& \texttt{CV-SVM} & 224 & 465 & 936 & 2364 & 4598 & 262 & 6597 & 659 & 108 & 229 & 443 & 1095 & 2070\\
& \subsvms & 4 & 5 & 5 & 6 & 7 & 5 & 8 & 4 & 4 & 4 & 5 & 6 & 7\\
\midrule
\multirow{5}{*}{$\alpha=0.0$} & \texttt{Oracle-SVM} & 118 & 180 & 256 & 393 & 598 & 31 & 257 & 28 & 120 & 181 & 237 & 347 & 495\\
& \texttt{Blind-SVM} & 118 & 171 & 240 & 385 & 571 & 31 & 273 & 22 & 123 & 176 & 229 & 329 & 482\\
& \texttt{Bag-SVM} & 21 & 42 & 80 & 197 & 342 & 46 & 899 & 105 & 14 & 29 & 54 & 112 & 206\\
& \texttt{CV-SVM} & 75 & 154 & 311 & 771 & 1510 & 141 & 2332 & 160 & 34 & 65 & 123 & 278 & 507\\
& \subsvms & 3 & 3 & 4 & 4 & 5 & 5 & 7 & 3 & 3 & 7 & 3 & 4 & 5\\
\bottomrule
\end{tabular}
\caption{Training times in seconds (rounded to the closest integer) for all the methods trained using L-2 loss averaged over 10 random label-manipulated versions of the data sets, corresponding to the results presented in Table 1 of the paper. `splc', `mush' and `svm1' stand for splice, mushrooms and svmguide1. Note that for \subsvms, the reported time is the time taken to train all the $J=1000$ SVMs on $s=\log^2\ell$-size subsets.}
\label{table:L2DetailedRunTimesBAC}
\end{table*}

\vskip 0.2in
\bibliography{srivats}


\end{document}